\documentclass[a4paper,fleqn]{cas-sc}

\usepackage[numbers,sort&compress]{natbib}

\usepackage{amsmath}
\usepackage{subfigure}
\usepackage{wrapfig}
\usepackage{array}
\usepackage{epstopdf}
\usepackage{float}
\usepackage{cases}
\usepackage{pifont}
\usepackage{amsfonts}
\usepackage{graphicx}
\usepackage{amssymb}
\usepackage{amstext}
\usepackage{enumerate}
\usepackage{color}

\usepackage{xcolor}
\usepackage{xpatch}
\usepackage{filecontents}

\usepackage{booktabs}
\usepackage{siunitx}
\usepackage{multirow}
\usepackage{enumitem}
\usepackage[linesnumbered,ruled,vlined]{algorithm2e}
\usepackage{hyperref}
\usepackage{tikz}
\newcommand*\circled[1]{\tikz[baseline=(char.base)]{
            \node[shape=circle,draw,inner sep=0.5pt] (char) {#1};}}
\def\tsc#1{\csdef{#1}{\textsc{\lowercase{#1}}\xspace}}
\tsc{WGM}
\tsc{QE}

\begin{document}
\let\WriteBookmarks\relax
\def\floatpagepagefraction{1}
\def\textpagefraction{.001}
\shorttitle{\underline{\leftline{\emph{D-DEPSO for Energy Saving FJSP Considering Machine Multi States}}}}
\shortauthors{D. Wang et al.}
\title [mode = title]{Discrete Differential Evolution Particle Swarm Optimization Algorithm for Energy Saving Flexible Job Shop Scheduling Problem Considering Machine Multi States}
\tnotemark[1]
\author[1]{Da Wang}
\credit{Conceptualization, Methodology, Supervision}
\author[1]{Yu Zhang}
\cormark[1]
\ead{2022021068@stu.sdnu.edu.cn}
\credit{Conceptualization, Writting, Software}
\author[1]{Kai Zhang}
\credit{Writting, Software, Editing}
\affiliation[1]{organization={Business School, Shandong Normal University},
            city={Ji'Nan},
            postcode={250014},
            state={Shandong Province},
            country={P.R.China}}

\author[2]{Junqing Li}
\credit{Supervision, Review, Technical Support}
\affiliation[2]{organization={School of  Mathematics, Yunnan  Normal University},
            city={Kun'Ming},
            postcode={650504},
            state={Yunnan Province},
            country={P.R.China}}
\author[3]{Dengwang Li}
\credit{Review, Editing, Technical Support}
\affiliation[3]{organization={School of Physics and Electronics, Shandong Normal University},
            city={Ji'Nan},
            postcode={250014},
            state={Shandong Province},
            country={P.R.China}}

\begin{abstract}
As the continuous deepening of low-carbon emission reduction policies, the manufacturing industries urgently need sensible energy-saving scheduling schemes to achieve the balance between improving production efficiency and reducing energy consumption. In energy-saving scheduling, reasonable machine states-switching is a key point to achieve expected goals, i.e., whether the machines need to switch speed between different operations, and whether the machines need to add extra setup time between different jobs. Regarding this matter, this work proposes a novel machine multi states-based energy saving flexible job scheduling problem (EFJSP-M), which simultaneously takes into account machine multi speeds and setup time. To address the proposed EFJSP-M, a kind of discrete differential evolution particle swarm optimization algorithm (D-DEPSO) is designed. In specific, D-DEPSO includes a hybrid initialization strategy to improve the initial population performance, an updating mechanism embedded with differential evolution operators to enhance population diversity, and a critical path variable neighborhood search strategy to expand the solution space. At last, based on datasets DPs and MKs, the experiment results compared with five state-of-the-art algorithms demonstrate the feasible of EFJSP-M and the superior of D-DEPSO.
\end{abstract}


\let\printorcid\relax
\maketitle

\section{Introduction}
\label{introduction}
With the advancement of global industrialization and the deepening of carbon reduction policies, traditional manufacturers are facing increasing challenges. How to effectively reduce energy consumption while pursuing production efficiency has become an urgent issue \cite{Destouet2023FlexibleJS,Xiong2022ASO}. Considering this, energy saving flexible job shop scheduling problem (EFJSP), which could provide theoretical assistant for manufacture managers to formulate more reasonable production plans, has emerged and gradually attracted widespread research interest \cite{Xu2021AMS,Wang2021EvolutionaryGB,Golmohammadi2015ASO,Song2023FlexibleJS,Park2022EnergyawareFJ}.

In EFJSP, the states of machines are constantly changing at different stages, among the cases of during, before, and after the operation process. On the one hand, during the operation process, each machine may have different speeds. Determining each operation’s speed and transporting different speeds between operations will both affect the time and energy consumptions \cite{yao1995scheduling,Lei2016ASF,Zhang2017MathematicalMA,Teymourifar2018ExtractingND,
Luo2019EnergyefficientSF,Wei2022HybridES,Lu2022AKM,Chen2023AHE,Meng2023MILPMA,zhuang2024multi}. On the other hand, before or after a process, each new job may need some extra setup time, encompassing tool changes, equipment cleaning, material loading and so forth \cite{Nouri2017SolvingTF,Shen2018SolvingTF,Sun2021AHM,Meng2022AnMM, Brandimarte2024ARV,Du2024ARL,Barak2024DualRC}. In a word, both multi speeds and extra setup time will make the machine multi states more complex and increase the difficulty of energy consumption calculation.

As for the multi speeds case, the earliest study can be found in \cite{yao1995scheduling}, in which the authors considered energy consumption issue of a power supply system with machine variable speeds. Recently, much more complicated situations considering multi speeds have attracted scholar's attention. For instance, Lei et al. \cite{Lei2016ASF} addressed the machine multi speeds EFJSP with the goal of workload balancing and the total energy consumption minimizing. Regarding machine selection, job sorting, and machine switching decisions, the work \cite{Zhang2017MathematicalMA} proposed a new class of FJSP model and provided decision lower bounds. Considering different job arrival times as well as the possibility of machine failures, some scheduling rules for general dynamic job-shop scheduling problems were investigated in \cite{Teymourifar2018ExtractingND}. According to the characteristics of multi-objective EFJSP, Luo et al. \cite{Luo2019EnergyefficientSF} constructed a bi-objective mathematical model based on discrete machine process speeds, and proposed a multi-objective grey wolf optimization algorithm to solve the model. A new energy-aware estimation model based on the variable speeds was formulated to calculate the energy consumption in different machine states \cite{Wei2022HybridES}. In \cite{Lu2022AKM}, the authors considered a class of green job-shop scheduling problems with variable process speeds (JSPVMS) and developed a new knowledge-based multi-objective memetic algorithm (MOMA) to address this problem. Based on the uncertainty of process time, a kind of fuzzy FJSP with variable process speed was studied in \cite{Chen2023AHE}, in which the fuzzy makespan and fuzzy total energy consumption were taken as objectives. Meng et al. \cite{Meng2023MILPMA} focused on the FJSP with controllable process time, used the exact algorithm to solve small-scale examples and designed a multi-objective shuffled frog leaping algorithm (MOSFLA) to solve middle and large scale problems. As introduced in \cite{zhuang2024multi}, the authors studied the impact of different machine speeds on makespan, energy consumption and tool wear, designed a three-layer encoding method including speed selection.

As for the setup time case, much attention has also been paid by scholars in consideration of different actual situations. Nouri et al. \cite{Nouri2017SolvingTF} constructed a hybrid meta heuristics-based clustered holonic multi-agent model, which was solved using a hybrid tabu search genetic algorithm. In \cite{Shen2018SolvingTF}, the FJSP with sequence setup time was studied to minimize the makespan. The authors used disjunctive graph model to analyze the problem structure and proposed a tabu search algorithm with specific neighborhood function and diversified structure. Based on a combination of setup time and transportation time, paper \cite{Sun2021AHM} simultaneously studied the four-objective FJSP including completion time, total workload, critical equipment workload, and earliness/tardy penalties. Meng et al. \cite{Meng2022AnMM} established a mixed integer linear programming model to solve the minimum energy consumption, analyzed the various energy consumption during the machine operation in detail, and carried out simulation experiments. A just-in-time job shop scheduling problem was developed in \cite{Brandimarte2024ARV}, in which the authors applied reduced variable neighborhood search method to deal with the sequence-dependent setup time situations. A kind of EFJSP considering crane transportation and installation time was studied in \cite{Du2024ARL}, in which the authors divided eight different crane transportation stages and three typical machine states. As introduced in \cite{Barak2024DualRC}, the authors constructed a three-objective model for the dual-resource constrained FJSP problem with sequence-dependent setup time, where the objectives include maximizing machine workload, minimizing personnel idle cost and setup cost.

The aforementioned investigations about EFJSP suggest that the two factors both have important influence on machine multi states, i.e., multi speeds and setup time. Nevertheless, to our best knowledge, few research has focused on the topic that considers the two aspects simultaneously. Therefore, there remains a need to further promote the EFJSP research by integrating both multi speeds and setup time.

Due to EFJSP is a kind of NP hard problem \cite{1009117}, it is difficult to obtain the exact solution and requires the assistance of optimization algorithms to find the optimal solution. It is worth noting that the numerical solution of EFJSP has obvious discrete characteristics, so lots of current researches have focused on algorithms with discrete evolution mechanisms \cite{Zhou2009MinimizingWT,Yazdani2019EvolutionaryAF,Li2021AnEM,Cheng2022SchedulingFM,Li2022AHI,Wang2022SolvingMF,Xie2023AHG,Wang2024AFL,Wei2024AnIM,Yan2024ALB}. Zhou et al. \cite{Zhou2009MinimizingWT} employed a hybrid framework combining a heuristic and a genetic algorithm (GA) to address the job-shop scheduling problem with minimum weighted tardiness objective. Two kinds of genetic algorithms were presented in \cite{Yazdani2019EvolutionaryAF} to address resource constrained FJSP problem. In \cite{Cheng2022SchedulingFM}, the authors proposed a Q-learning based genetic algorithm to solve large-scale problems in flow production and job production scheduling. By combining iterative greedy and simulated annealing algorithms to devise a tailored heuristic approach, the authors \cite{Li2022AHI} investigated the EFJSP involving crane transportation. Xie et al. \cite{Xie2023AHG} integrated the global search performance of the genetic algorithm with the local search precision of tabu search to solve the dynamic FJSP. As introduced in \cite{Wang2024AFL}, a feedback learning-based memetic algorithm was presented to address the cooperative scheduling in distributed job shops with limited transportation resources. Considering the uncertainty scheduling problem caused by resource constraints, the study \cite{Wei2024AnIM} designed a four objectives optimization model and developed an improved meme algorithm to solve it. By introducing the dual population idea, Yan et al. \cite{Yan2024ALB} designed an auxiliary dual population evolutionary algorithm to solve multi-objective FJSP with project management. In addition, given the characteristics of few parameters and easy implementation, algorithms with continuous evolution mechanisms such as PSO have also attracted scholars' attention to solve EFJSP \cite{zhou2020particle,Tan2021LowcarbonJS,Fontes2023AHP,Shi2023ANB}. For instance, Zhou et al. \cite{zhou2020particle} incorporated PSO into the multi-objective evolutionary algorithm, and proposed a hybrid algorithm containing particle filter and lévy flight to tackle enhanced EFJSP. In \cite{Tan2021LowcarbonJS}, the authors designed an improved multi-objective PSO that enhances convergence through refined initialization, chaotic position updating, and mutation strategies. Fontes et al. \cite{Fontes2023AHP} presented a hybrid approach combining PSO and simulated annealing (PSOSA), leveraging it to develop a swift lower bound process for solving FJSP. As presented in \cite{Shi2023ANB}, a two-stage multi-objective PSO algorithm (MOPSO) based on a three-dimensional representation was investigated to deal with FJSP considering worker efficiency. In summary, both continuous and discrete algorithms have their own characteristics and advantages, therefore, this work aims to fully combine the advantages of them. In specific, this work intends to embed the idea of differential evolution into the discrete PSO framework to address EFJSP.

Based on the above discussions, the purpose of this work is to construct a kind of EFJSP considering machine multi states, and propose a discrete differential evolution particle swarm optimization algorithm to solve it. Specifically, the contributions of this paper are as follows:

\begin{enumerate}
\item[(1)] Construct a kind of energy saving flexible job shop scheduling problem considering machine multi states (EFJSP-M). The constructed EFJSP-M is more in line with the real production characteristics, which involves both setup time and multi process speeds.
\item[(2)] Propose a discrete particle swarm algorithm embedded with differential evolution (D-DEPSO). The updating scheme of D-DEPSO makes full use of two algorithms’ advantages and improves the optimization performance.
\item[(3)] In D-DEPSO, a particle initialization method based on random and heuristic rules is presented. The hybrid initialization mechanism is benefit to reduce the blindness of particle search.
\end{enumerate}

The outline of this study is given as follows. Section \ref{Model} gives the problem statement, symbol representation and model building process of EFJSP-M. In Section \ref{Algorithm}, the basic framework of D-DEPSO is introduced, including coding and decoding, population initialization strategy, updating mechanism and variable neighborhood search. The simulation experiment and result analysis are given in Section \ref{Experiment}, which includes model test, parameter experiment, comparison analysis with algorithm variations, and comparison analysis with five other algorithms. In Section \ref{Conclusion}, the summary of this work and future research indications are made.

\section{Model construction}
\label{Model}
\subsection{Problem statement}
\label{2.1}
\vspace{0.2cm}
In the proposed EFJSP-M, there are $n$ independent jobs to be executed on $k$ machines in a workshop. The job set and the machine set are denoted as $I=\{1,2,\ldots,n\}$, $M=\{1,2,\ldots,k\}$, respectively. Each job $i$ contains several operations, where $N_i=\{1,2,\ldots,n_i\}$ represents the operations set for job $i$ and $n_i$ represents the operations amount for job $i$. Each operation can select one machine from its candidate set for process. Each machine has multi speeds  $V=\{1,2,\ldots,s\}$ to choose from, while the process time and energy consumption are different at different speeds. Each operation has a different standard process time on each machine, and the process time varies according to the speeds. The machine has four operating states (setup, process, idle, and standby) and six state switching commands (turn on, dormancy, dormancy-release, deceleration, acceleration, and turn off).

A reasonable scheduling scheme is supposed to put forward to help enterprises carry out production planning and improve work efficiency. Subtasks of the scheduling problem include: (1) assigning a suitable machine and appropriate process speed for each operation; (2) sequencing operations for each machine.

\subsection{Notations}
\label{2.2}
\vspace{0.2cm}
The symbols used in the context of the EFJSP-M are defined in Table \ref{bianliang}.

\begin{table}[!h]
\centering
\caption{Relevant variable and meanings.}
\label{bianliang}
\vskip 3pt
\begin{tabular}{ll}
\hline
  Variable&Meaning \\
\hline
 $I$ &Set of jobs\\
 $N_i$ &Set of operations for job $i$\\
 $M$ &Set of machines\\
 $V$ &Set of speed\\
 $i,i'$ &Index of jobs\\
 $j,j'$ &Index of operations\\
 $m$ &Index of machines\\
 $v$ &Index of speed\\
 $t$ &Index of idle/standby interval\\
 $n$ &Number of jobs\\
 $n_i$ &Number of operations for job $i$\\
 $n_m$ &Number of operations for machine $m$\\
 $k$ &Number of machines\\
 $s$ &Number of speed\\
 $n_t$ &Number of idle/standby interval\\
 $E$ &Energy consumption\\
 $IE$ &Instruction energy consumption\\
 $SE$ &State energy consumption\\
 $P$ &Machine power\\
 $T$ &Running time\\
 $SU_i$ &Setup time of job $i$\\
 $J_i$ &Job $i$\\
 $O_{ij}$ &Operations $j$ for job $i$\\
 $S_{ij}$ &Start time of $O_{ij}$\\
 $C_{ij}$ &Completion time of $O_{ij}$\\
 $P_{ij}$ &Process time of $O_{ij}$\\
 $o$ &Speed 0\\
 $V_{mj}$ &The speed of operation $j$ on machine $m$\\
 $V_t$ &The speed of idle interval $t$\\
 $V_0$ &The speed of standby state\\
\hline
\end{tabular}
\end{table}

\subsection{Assumptions}
\label{2.3}
\vspace{0.2cm}
In this paper, simplified assumptions are set according to the actual production situations:

\begin{enumerate}
\item[(1)] The sequence of operations for each job is fixed.

\item[(2)] Only one machine can be selected from the available machine set for each operation at a time.

\item[(3)] Only one operation can be processed for each machine at a time.

\item[(4)] The operation process does not allow interruption.

\item[(5)] All machines and operations are available at time 0.

\item[(6)] Once the operation starts process, the speed is not allowed to change.

\item[(7)] The buffer is infinite, i.e., the machine can enter the next state immediately after the current operation is processed.

\end{enumerate}

\subsection{Mathematic model of EFJSP-M}
\label{2.4}
\vspace{0.2cm}
To present the model more intuitively, it is necessary to analyze the multi energy consumption in different cases. In specific, energy consumption occurs in the following two situations: during a machine state ($S$), and executing states switching instructions ($I$).

First of all, once the machine starts at time 0, the instruction energy consumption is generated immediately, which is denoted as the turn-on energy consumption ${IE}_1$. Considering that different operations in the actual production process need a certain setup time before processing, such as cleaning, reconfiguration, etc., the machine should first enter the setup state when switching to an operation of a new job. In this case, instruction energy consumption and state energy consumption will both be generated, which are denoted as acceleration/deceleration energy consumption ${IE}_2$ and setup energy consumption $SE_1$, respectively. It is worth noting that if two adjacent operations on the same machine belong to the same job, there is no need to enter the setup state again. After the setup state, the machine immediately enters the process state, and the process energy consumption is generated, denoted as ${SE}_2$. Then, after process state is completed, the machine will choose to enter the idle state or standby state, and the corresponding energy consumption is simplified as $ISE$ (the detailed presentation of $ISE$ can be seen in (\ref{ISE})). After all operations are completed, the machine executes the turn-off instruction, which has no energy consumption.

To sum up, the total energy consumption $TEC$ of the machine is composed of five parts, and the calculation formula is shown as:
\begin{equation}\label{zongnenghao}
  TEC = {IE}_1 + {IE}_2 + {SE}_1 + {SE}_2 + ISE.
\end{equation}

All machines will generate turn-on energy consumption when starting up, and the machine will accelerate to the speed of the first operation. Thus, the turn-on energy consumptions are different according to different starting speeds. Therefore, for ${IE}_1$, we should determine the speed of first operation on all machines and calculate the corresponding turn-on energy consumption. The calculation formula of $IE_1$ is shown as:
\begin{equation}\label{IE1}
  {IE}_1 = \sum_{m=1}^{k} E_{o\rightarrow v_{m1}}.
\end{equation}

${IE}_2$ only exists between continuous operations (the machine does not enter the idle or standby state between the two operations), thus we just need find all continuous operations in the scheduling table and calculate the corresponding energy consumption. The calculation formula of $IE_2$ is shown as:
\begin{equation}\label{IE2}
  {IE}_2 = \sum_{m=1}^{k} E_{v_{mj}\rightarrow v_{mj+1}}\cdot Z_{mjj+1},j=1,2,\ldots, n_m.
\end{equation}

The setup power of different machines is different, at the same time, the setup time of different jobs is different. Therefore, the calculation method of $SE_1$ can be summarized as follows. Before each operation is added to the scheduling plan, all idle time periods on the selected machine are traversed. The setup time needs to be added if one of the following two cases is satisfied: (1) no operation is scheduled before the idle time period, (2) the previous operation of the idle time period is not the same job as the current operation. The calculation formula of $SE_1$ is shown as:
\begin{equation}\label{SE1}
  {SE}_1 = \sum_{m=1}^{k}\sum_{i=1}^{n}\sum_{j=1}^{n_i} P_1\cdot T_{1i}\cdot X_{ijm}\cdot (1-Y_{mij-1j}).
\end{equation}

Process energy consumption is equal to the product of process time and process power. Assume that there is an standard process power for a machine, the actual process power varies at different process speeds. The calculation formula of $SE_2$ is shown as:
\begin{equation}\label{SE2}
  {SE}_2 = \sum_{i=1}^{n}\sum_{j=1}^{n_i}\sum_{m=1}^{k}\sum_{v=1}^{s} P_{2v}\cdot T_{2ijkv}\cdot X_{ijmv}.
\end{equation}

As for $ISE$, first calculate the energy consumption in idle state and standby state respectively, then take the smaller one. Among them, the energy consumption of former idle state includes the state energy consumption of maintaining idle state at a lower rate and the instruction energy consumption of primary speed switching command. The energy consumption of the latter standby state includes the energy consumption of maintaining the standby state and the instruction energy consumption of two speed switching commands. The calculation formulas of $ISE$ are shown as:
\begin{equation}\label{ISE}
  ISE = min(E_{idle}, E_{standby}).
\end{equation}
\begin{equation}\label{Eidle}
  E_{idle} = \sum_{m=1}^{k}\sum_{t=1}^{n_t}P_{3v_t}\cdot T_{3t} + \sum_{m=1}^{k}E_{v_{mj}\rightarrow v_t}\cdot \theta + \sum_{m=1}^{k}E_{v_t\rightarrow v_{mj+1}}\cdot (1 - \theta),j = 1,2,\ldots,n_m.
\end{equation}
\begin{equation}\label{Estandby}
  E_{standby} = \sum_{m=1}^{k}\sum_{t=1}^{n_t}P_{4v_0}\cdot T_{4t} + \sum_{m=1}^{k}E_{v_{mj}\rightarrow v_0} + \sum_{m=1}^{k}E_{v_0\rightarrow v_{mj+1}},j = 1,2,\ldots,n_m.
\end{equation}

Based on the above analysis, the energy-saving scheduling model considering machine multi states is given as follows:

\textbf{Objectives:}
\begin{equation}\label{objectives}
\begin{aligned}
  minf_1 = C_{max}. \\
  minf_2 = TEC.
\end{aligned}
\end{equation}

\textbf{Subject to:}
\begin{equation}\label{eq1}
  C_{max}\geq C_{in_i},\forall i\in I.
\end{equation}
\begin{equation}\label{eq2}
  S_{ij}+P_{ij}=C_{ij},\forall i \in I, j\in N_i.
\end{equation}
\begin{equation}\label{eq3}
  C_{ij} \leq S_{ij+1},\forall i \in I, j=1,2,\ldots,n_i-1.
\end{equation}
\begin{equation}\label{eq4}
  y_{ijm} \leq a_{ijm},\forall i \in I, j\in N_i,m \in M.
\end{equation}
\begin{equation}\label{eq5}
  \sum_{m=1}^{k}y_{ijm}=1,\forall i \in I, j\in N_i.
\end{equation}
\begin{equation}\label{eq6}
  \sum_{v=1}^{s}z_{ijv}=1,\forall i \in I, j\in N_i.
\end{equation}
\begin{equation}\label{eq7}
  {yz}_{ijmv} \leq y_{ijm},\forall i \in I, j\in N_i,m \in M.
\end{equation}
\begin{equation}\label{eq8}
  {yz}_{ijmv} \leq z_{ijv},\forall i \in I, j\in N_i,v \in V.
\end{equation}
\begin{equation}\label{eq9}
  \sum_{m=1}^{k}\sum_{v=1}^{s} {yz}_{ijmv}=1,\forall i \in I, j\in N_i.
\end{equation}
\begin{equation}\label{eq10}
  P_{ij}=\sum_{m=1}^{k}\sum_{v=1}^{s} {yz}_{ijmv}\cdot p_{ijmv},\forall i \in I, j\in N_i.
\end{equation}
\begin{equation}\label{eq11}
  \sum_{i=1}^{n}\sum_{j=1}^{n_i}x_{m00ij}=1,\forall m \in M.
\end{equation}
\begin{equation}\label{eq12}
  \sum_{i=0}^{n}\sum_{j=0}^{n_i}x_{miji'j'}=y_{i'j'm},\forall i' \in I, j'\in N_i,m \in M.
\end{equation}
\begin{equation}\label{eq13}
  \sum_{i'=0}^{n}\sum_{j'=0}^{n_i}x_{miji'j'}=y_{ijm},\forall i \in I, j\in N_i,m \in M.
\end{equation}
\begin{equation}\label{eq14}
  x_{miji'j'}+x_{mi'j'ij} \leq 1,\forall i,i'\in I, j\in N_i, j' \in N_{i'},m \in M.
\end{equation}
\begin{equation}\label{eq15}
  x_{mijij}=0,\forall i \in I, j\in N_i,m \in M.
\end{equation}
\begin{equation}\label{eq16}
  x_{mi0i'j'}=0,\forall i,i'\in I,i'\neq i, j' \in N_{i'},m \in M.
\end{equation}
\begin{equation}\label{eq17}
  x_{miji'0}=0,\forall i,i'\in I,i'\neq i, j\in N_i,m \in M.
\end{equation}
\begin{equation}\label{eq18}
  S_{ij}+P_{ij} \leq S_{i'j'}+M\cdot(1-x_{miji'j'}),\forall i,i'\in I, j\in N_i, j' \in N_{i'},m \in M.
\end{equation}
\begin{equation}\label{eq19}
  S_{ij} \geq {SU}_i-M\cdot(1-x_{m00i'j'}),\forall i, j\in N_i,m \in M.
\end{equation}
\begin{equation}\label{eq20}
  S_{i'j'} \geq C_{ij}-M \cdot (1-x_{miji'j'}),\forall i,i'\in I, j\in N_i, j' \in N_{i'},m \in M.
\end{equation}
\begin{equation}\label{eq21}
  a_{ijm}=\left\{\begin{array}{l}
  1, if~O_{ij}~can~be~processed~on~machine~m, \\
  0, otherwise.\\
  \end{array}\right.
\end{equation}
\begin{equation}\label{eq22}
  y_{ijm}=\left\{\begin{array}{l}
  1, if~O_{ij}~is~selected~to~be~processed~on~machine~m, \\
  0, otherwise.
  \end{array}\right.
\end{equation}
\begin{equation}\label{eq23}
  z_{ijm}=\left\{\begin{array}{l}
  1, if~O_{ij}~is~selected~to~be~processed~by~speed~v, \\
  0, otherwise.
  \end{array}\right.
\end{equation}
\begin{equation}\label{eq24}
  p_{ijmv}=\left\{\begin{array}{l}
  1, if~O_{ij}~is~selected~to~be~processed~on~machine~m~by~speed~v, \\
  0, otherwise.
  \end{array}\right.
\end{equation}
\begin{equation}\label{eq25}
  x_{miji'j'}=\left\{\begin{array}{l}
  1, if~O_{ij}~and~O_{i'j'}~is~selected~to~be~processed~on~machine~m~and~O_{ij}~is~the~direct~ precursor~to~O_{i'j'}, \\
  0, otherwise.
  \end{array}\right.
\end{equation}

As shown in formula (\ref{objectives}), the goal of the model is to minimize the maximum completion time $C_{max}$ and total energy consumption $TEC$. Among them, $C_{max}$ is the latest completion time for all operations, as shown in constraint (\ref{eq1}). Constraint (\ref{eq2}) indicates that the processing of each operation is not allowed to be interrupted. Constraint (\ref{eq3}) ensures the operations' priorities of the same job. Constraint (\ref{eq4}) specifies that each operation can only select machines in its optional machine set. Constraints (\ref{eq5}-\ref{eq9}) ensure that each operation must select and can only select one machine speed combination. Constraint (\ref{eq10}) determines the corresponding process time after each operation's machine and speed are determined. Constraint (\ref{eq11}) guarantees that in the first place, each machine has only one, and must have one operation. Constraints (\ref{eq12}-\ref{eq17}) give priority constraints between adjacent operations. Constraint (\ref{eq18}) indicates that if $O_{ij}$ is  preceding operation of $O_{i'j'}$, then $O_{i'j'}$ must begin processing after $O_{ij}$ is completed. Constraints (\ref{eq19}-\ref{eq20}) add setup time to the model. Constraints (\ref{eq21}-\ref{eq25}) give the specific meaning of decision variables.

\subsection{Model sample}
\label{2.5}
\vspace{0.2cm}
In order to better explain the model, a detailed example is designed in Table \ref{jiagongnenghao} and Table \ref{qitanenghao}. The example contains two jobs and two machines, job 1 has two operations, job 2 has four operations. Table \ref{jiagongnenghao} shows the optional process machines for each operation and the process time at different speeds. The symbol ‘-‘ indicates that the machine is not available for this operation. The last column shows the setup time of the two jobs. Table \ref{qitanenghao} illustrates the energy consumptions of executing speed switching instructions $IE$ and non-processing state energy consumption at different speeds $SE$.

\begin{table}[!h]
\centering
\caption{Process time and setup time of the sample.}
\label{jiagongnenghao}
\vskip 3pt
\begin{tabular}{ccccccccc}
\hline
\multirow{2}{*}{$Job_{ID}$} & \multirow{2}{*}{$Operation_{ID}$} &
\multicolumn{3}{c}{$M_1$}&\multicolumn{3}{c}{$M_2$} &
\multirow{2}{*}{$SU$}\\ \cline{3-8} &
 & $v_1$ & $v_2$ & $v_3$ & $v_1$ & $v_2$ & $v_3$ \\ \hline
\multirow{2}{*}{$J_1$} &
 $O_{11}$ & 18 & 12 & \textbf{6} & - & - & - &
\multirow{2}{*}{1}\\ &
 $O_{12}$ & 6 & 4 & \textbf{2} & - & - & - &\\ \hline
 \multirow{4}{*}{$J_2$} &
 $O_{21}$ & 30 & 20 & 10 & 27 & 18 & \textbf{9} &
\multirow{4}{*}{2}\\ &
 $O_{22}$ & 4 & 2 & 1 & 6 & \textbf{4} & 2 &\\
 & $O_{23}$ & 6 & \textbf{3} & 1 & - & - & - &\\
 & $O_{24}$ & - & - & - & 9 & 6 & \textbf{3} &\\
\hline
\end{tabular}
\end{table}

\begin{table}[!h]
\centering
\caption{Speed switching instruction energy consumption and idle/standby state energy consumption of the sample.}
\label{qitanenghao}
\vskip 3pt
\begin{tabular}{ccccc}
\hline
  $IE$ & $v_0$ & $v_1$ & $v_2$ & $v_3$\\
\hline
 $v_0$ & 0 & 5 & 8 & 10\\
 $v_1$ & 5 & 0 & 5 & 8\\
 $v_2$ & 8 & 5 & 0 & 5\\
 $v_3$ & 10 & 8 & 5 & 0\\
\hline
$SE$ & 2 & 3 & 6 & 9\\
\hline
\end{tabular}
\end{table}

\begin{figure}[h]
\begin {center}
\includegraphics[height=1.8in,width=3.6in]{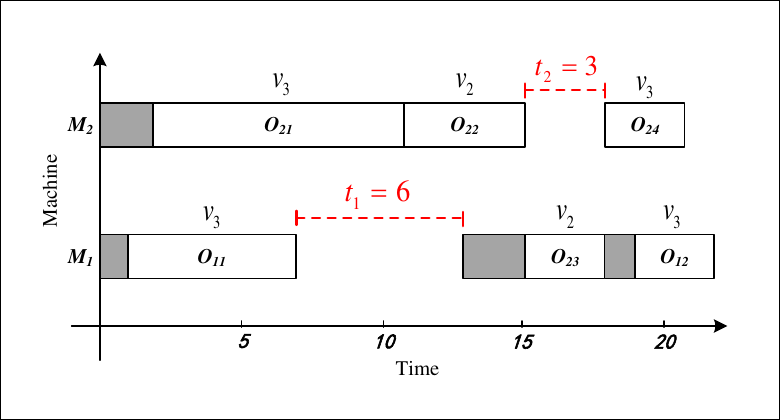}
\caption {The Gantt Diagram of the sample.}
 \label{Fig.example1} 
 \end{center}
\end{figure}

In Fig.\ref{Fig.example1},  we present the Gantt Diagram of a feasible solution with the process sequence $O_{11} \rightarrow O_{21} \rightarrow O_{22} \rightarrow O_{23} \rightarrow O_{24} \rightarrow O_{12}$. The machine and speed selections of each operation are marked in bold in Table \ref{jiagongnenghao}. One can see from Fig.\ref{Fig.example1} that $O_{11}$ is processed on $M_1$ at $v_3$. In this case, $O_{11}$ has setup time $1$ and process time $6$. The operation $O_{21}$ on machine $M_2$ at speed $v_3$ needs setup time $2$ and process time $9$. Note that $O_{22}$ is processed on $M_2$, on which the direct precursor $O_{21}$ also belongs to job $J_2$, so there is no need to setup for $O_{22}$. That is, $O_{22}$ could start at time $11$, after its process time $4$, end at time $15$. As for $O_{23}$, a new setup time is needed since it switches to machine $M_1$. This setup can be completed before time node 15, such that $O_{23}$ can start smoothly after $O_{22}$. The subsequent operations are arranged in the same way as above.

\begin{figure}[h]
\begin {center}
\includegraphics[height=3.7in]{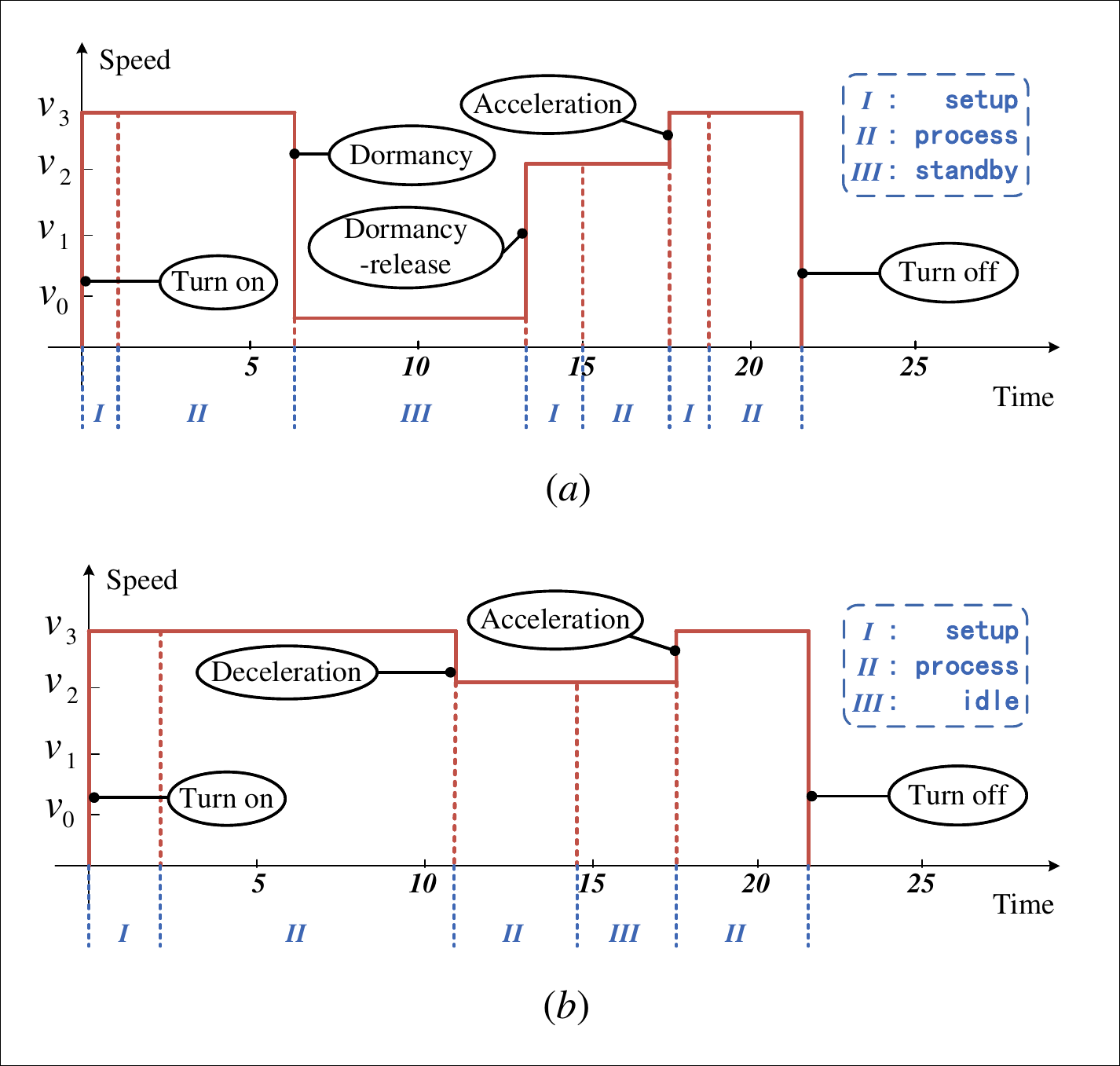}
\caption {The sample's machine states during the whole cycle:(a)$M_1$,(b)$M_2$.}
 \label{Fig.example2} 
 \end{center}
\end{figure}

It is worth noting that $M_1$ has an interval between $O_{11}$ and $O_{23}$, denoted as ${interval}_1$ with $t_1=6$. Then we will judge ${interval}_1$'s machine running state, idle or standby, according to the rules in formulas (\ref{Eidle}) and (\ref{Estandby}). If choose idle state for ${interval}_1$, the machine should change from $v_3$ to $v_2$, with the result that $E_{idle}=6\times6+5=41$. If choose standby state for ${interval}_1$, the machine speed should firstly switch from $v_3$ to $v_0$, then change from $v_0$ to $v_2$ when the next operation $O_{23}$ begins. That is, $E_{standby}=2\times6+10+8=30$. Since $E_{standby}<E_{idle}$, we choose ${ISE}_{{interval}_1}= E_{standby}=30$. So, $M_1$ runs standby in ${interval}_1$ at $v_2$. Similarly, $M_2$ has an interval between $O_{22}$ and $O_{24}$, denoted as ${interval}_2$ with time 3. In ${interval}_2$, $E_{idle}=6\times3+5=23$, $E_{standby}=2\times3+8+10=24$. Since $E_{idle}<E_{standby}$, we choose ${ISE}_{{interval}_1}= E_{idle}=23$. Therefore, $M_2$ runs idle in ${interval}_2$ at $v_2$. Based on the above analysis, the detailed running trajectories of $M_1$ and $M_2$ are listed in Fig. \ref{Fig.example2}. For instance, at time node between stage II and stage III in Fig. \ref{Fig.example2}(a), machine $M_1$ should switch from speed $v_3$ to $v_0$.

\section{The proposed D-DEPSO}
\label{Algorithm}

\subsection{Main Framework}
\label{3.1}
\vspace{0.2cm}

 \begin{figure}[h]
\begin {center}
\includegraphics[height=5.7in]{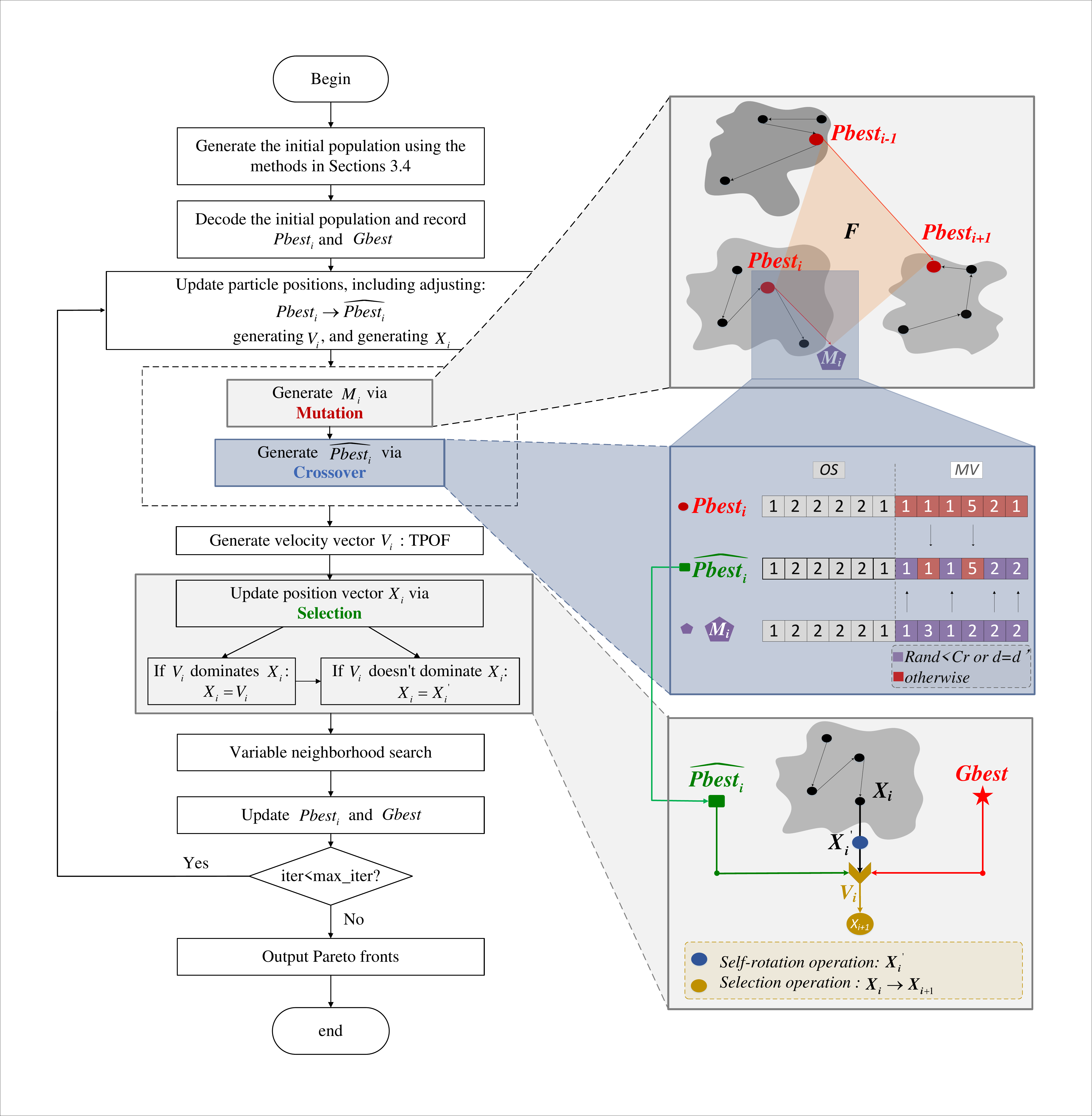}
\caption {The framework of D-DEPSO.}
 \label{Fig.flowchart}
 \end{center}
\end{figure}

PSO has gained popularity due to its ease of implementation, less parameters, and fast convergence speed. This work attempts to introduce PSO to address the above proposed EFJSP-M. Nevertheless, the canonical PSOs are more suitable for solving continuous optimization problems. When facing with discrete optimization problems whose solution is integers combination, PSO often leads to infeasible situations. Thus, we construct a kind of discrete version PSO, denoted as D-PSO, which contains discrete initialization scheme and modified updating mechanism. In addition, when solving complex optimization problems, PSO often suffers from reduced exploration ability and falls into local optima due to the decline of population diversity. Therefore, the idea of differential evolution (DE), well known for efficient exploitation capabilities, is embedded in D-PSO to further improve optimization performance. The framework of the proposed hybrid algorithm, denoted as D-DEPSO, is illustrated in Fig.\ref{Fig.flowchart}. In detail, the D-DEPSO contains the following six parts.

\begin{enumerate}[label=step\arabic*, ref=step\arabic*]
\item Encoding. A two-layer coding method is designed to represent job sequences and machine-speed combination.
\item Population initialization. A kind of hybrid initialization method is designed, which contains random and heuristic generation mechanism. The initialized particle position is denoted as $X_i^1$.
\item Decoding. A type of activity decoding method that considers setup time is proposed to translate the digital codes into actual schedule information, including machine IDs, speed selection, and job sequence. Record individual best-performed position and global optimization for the entire population as ${Pbest}_i$ and $Gbest$, respectively.
\item \label{step:4}Population evolution. Firstly, the mutation and crossover ideas in differential evolution are introduced to generate the modified adjust $Pbest_i$, which is denoted as $\widehat{Pbest_i}$. Secondly, design a noval TPOF operator, which could fuse  the information of three exemplars, i.e., $X_i$, $\widehat{Pbest_i}$ and $Gbest$, to generate new velocity vector $v_i$. Finally, update the particle position vector via the selection mechanism of differential evolution.
\item Variable neighborhood search. A variable neighborhood search based on the critical path is performed on the non-dominated solution generated by the current iteration.
\item Renewal ${Pbest}_i$ and $Gbest$. If the number of iterations reaches the predefined maximum number of iterations, the algorithm terminates and outputs the Pareto fronts. Otherwise, continue to \ref{step:4}.
\end{enumerate}

\subsection{Encoding scheme}
\label{3.2}
\vspace{0.2cm}
The PSO is initially designed for solving continuous optimization problems, while the scheduling problem represents a complex discrete optimization challenge. Therefore, it is crucial to devise an appropriate coding scheme that aligns PSO with the requirements of the scheduling problem. On the one hand, this coding scheme must ensure the feasibility of the algorithm's update process. On the other hand, it should possess a high level of scalability. Compared to the FJSP problem, EFJSP-M requires further selection of machine speed based on the determined machine ID. On the basis of the original machine selection (MS) + operation sequence (OS) structure, this paper extends MS to the combination of machine and speed (MV) with reference to \cite{Wei2022HybridES}.

\begin{figure}[h]
\begin {center}
\includegraphics[width=4.2in]{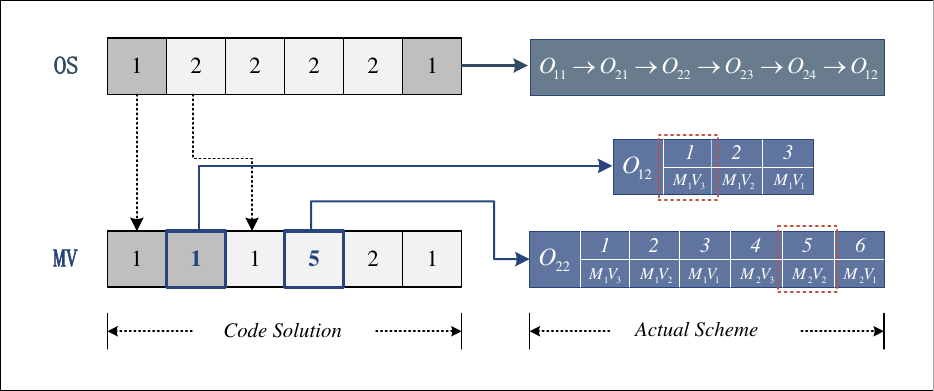}
\caption {The illustration of two-dimensional extended coding scheme.}
 \label{Fig.encoding}
 \end{center}
\end{figure}

As shown in Fig.\ref{Fig.encoding}, OS is a vector with the length of the total operations. In OS, each number itself represents the job id, and the frequency of the same number occurrences represents the operation id. Therefore, this encoding mechanism can easily represent the complete process sequence. In addition, it also has high scalability since the transposition of the point order will not cause infeasibility. As for MV, it is also a vector with the same length as OS. In MV, a number's position determines its corresponding job and operation, and the number itself represents the selection of machine-speed combination. The numerical exchange between the same positions of different MVs will not change their feasibility. Besides, in MV, replacing a number by its corresponding operation's other available machine-speed combination will not change the feasibility.

\subsection{Decoding scheme}
\label{3.3}
\vspace{0.2cm}
In this paper, an active decoding method considering setup time (ADM-S) is proposed. First, for an operation, one need to find an earliest available idle interval on the corresponding machine. Then, it is also necessary to consider whether to add a setup time before inserting the operation into this interval. In particular, Algorithm \ref{algorithm1} lists the pseudo-code of ADM-S. Its specific steps are described as follows:

\begin{itemize}
  \item \textbf{Preparations}
    \begin{enumerate}
    \item[(i)] Create an empty scheduling table, in which each row contains an operation's job id, operation id, machine number, speed gear, start time and end time.
    \item[(ii)] For each operation, establish a message matrix with three rows. The first row represents the optional machine id. The second row lists the available speeds. The third row gives the corresponding process time. In specific, the data in a message matrix is sorted in ascending order of the third row, that is, process time. For instance, the message matrix for $O_{22}$ of the sample data in Section \ref{2.5} is shown as follows:
$$Message~Matrix~_{O_{22}}=
\begin{bmatrix}
1&1&2&1&2&2\\3&2&3&1&2&1\\1&2&2&4&4&6
\end{bmatrix}.
$$

    \end{enumerate}
  \item \textbf{Particle information analysis}
    \begin{enumerate}
    \item[(iii)] For each operation in a particle's OS part, determine its chosen machine and process speed according to MV and the message matrix.
    \item[(iv)] Based on its process machine id, find out all the idle intervals of this machine in the scheduling table. Arrange the idle intervals in ascending order according to the start time.
    \end{enumerate}
  \item \textbf{Scheduling table update}
    \begin{enumerate}
    \item[(v)] For an operation to be filled in the table, iterate through the interval obtained in (iv). Determine whether the current operation needs to add the setup time according to the following two cases:\\
            \textbf{Case 1:}\label{case1} No job is scheduled before this interval, that is, the current operation is the first operation on the machine;\\
            \textbf{Case 2:}\label{case2} On the process machine, the operation to be inserted and its direct precursor belong to different jobs.
    \item[(vi)] Update table information. If meet the two cases above, add a setup record row in the scheduling table with the operation id is set to 0. That is, regard it as a virtual operation.
    \end{enumerate}
  \item \textbf{Fitness calculation}
    \begin{enumerate}
    \item[(vii)] Based on the final table, read the $C_{max}$, calculate the $TEC$. Complete a decoding process.
    \end{enumerate}
\end{itemize}

\begin{algorithm}[H]
\label{algorithm1}
 \SetAlgoLined
 \KwIn{Particle i}
 \KwOut{Schedule}
 Identify OS and MV\;
 \For{$O_{ij}~in~OS$}{
 Find the machine $M$ assigned to $O_{ij}$ in $MV$\;
 Get all scheduling data for $M$\;
 Calculate all interval data for $M$ and sort in ascending order by start time $\rightarrow Emptyslot$\;
 \For{$interval~in~Emptyslot$}{
 \If{Not(case 1 and case 2)}{
  Add setup operation\;
  Add process operation\;
  \If{the time constraint is met}{
  Do Add\;
  Break;
   }{Undo Add\;}
  }
  Add process operation\;
  \If{the time constraint is met}{
  Do Add\;
  Break;
   }{Undo Add\;}
  }
 }
 \caption{Process of ADM-S.}
\end{algorithm}

\begin{figure}[h]
\begin {center}
\includegraphics[height=3.5in]{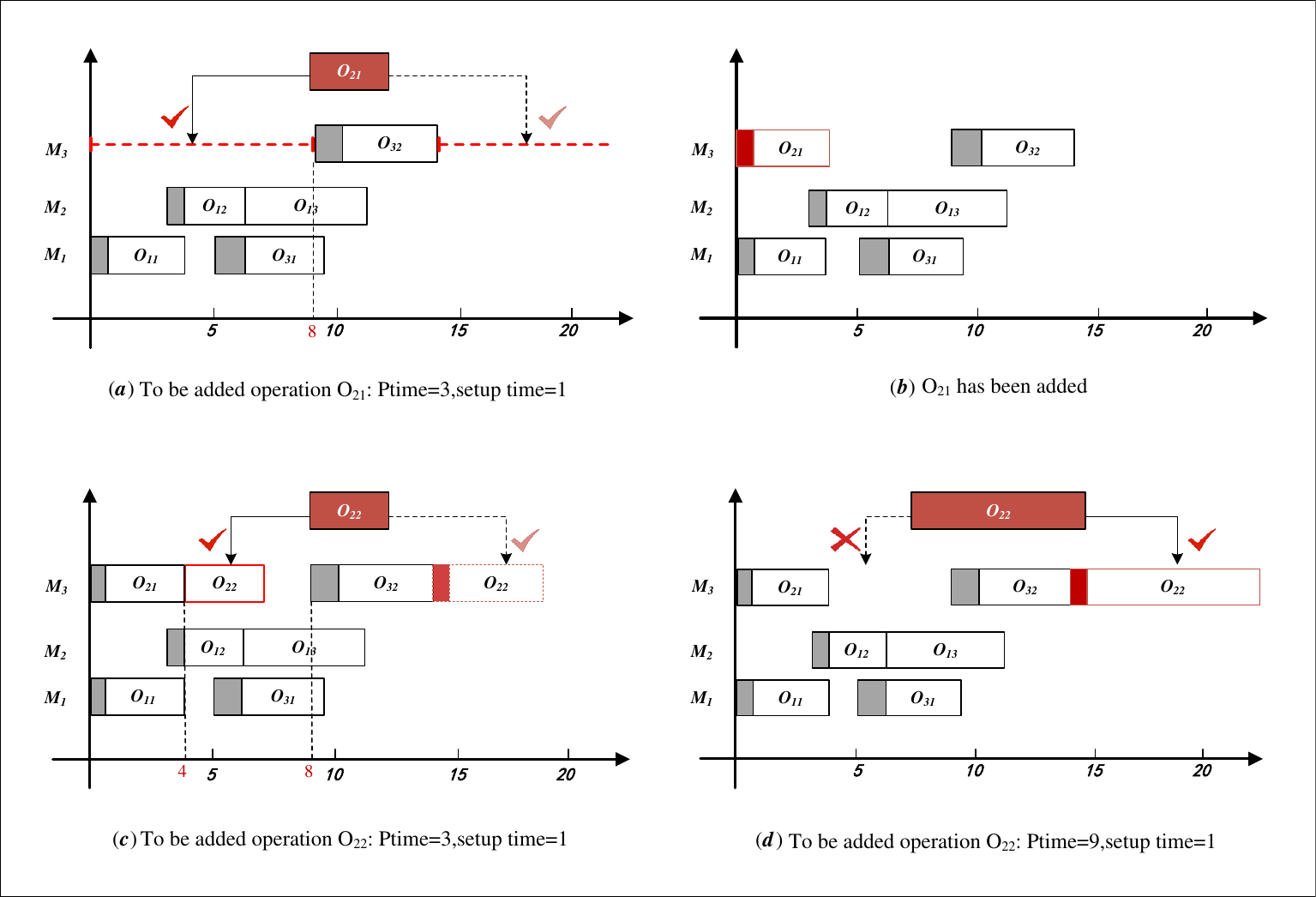}
\caption {The illustration of ADM-S.}
 \label{Fig.ADM_S}
 \end{center}
\end{figure}

To better demonstrate the aforementioned ADM-S, an example is shown in Fig.\ref{Fig.ADM_S}. Assuming that the machine id of $O_{21}$ and $O_{22}$ is $M_3$. Specifically, $O_{21}$ is the next operation to be processed into the current scheduled Gantt Diagram. As illustrated in Fig.\ref{Fig.ADM_S}(a), there are two intervals on machine $M_3$, that is, both sides of $O_{32}$. Based on (iv), we first determine whether the early started left interval has enough space to arrange $O_{21}$. Since Case 1 in (v) is satisfied, the total duration of $O_{21}$ is $process time + setup time = 3+1 = 4 < 8$. That is, $O_{21}$ will be inserted into the left interval. In Fig.\ref{Fig.ADM_S}(c), if $O_{22}$ is scheduled as a direct successor to $O_{21}$, there is no need to add setup time. Since $process time = 3 < 4$, this scheme is feasible. On the contrary, in Fig.\ref{Fig.ADM_S}(d), because the process time of $O_{22}$ is too long, it can only be arranged after $O_{32}$. In this way, Case 2 in (v) is satisfied, therefore the setup time needs to be added.

\subsection{Initialization method}
\label{3.4}
\vspace{0.2cm}
For particle swarm optimization, a good initialization method should ensure the high quality distribution of whole swarm in the search space. In specific, EFJSP-M is a global optimal problem that involves many individual operations. To some extent, the optimal solution definitely includes some individuals' own optimal fragments. To better utilize these fragments, this paper adopts a combination of random and heuristic initialization method to improve population quality. In this combination initialization method, all OS are generated randomly. As for MV, there are two heuristic rules:

 \textbf{Rule 1:}\label{Rule 1} If each operation can achieve the shortest process time, the scheduling is more likely to be completed in a relatively short time.

 \textbf{Rule 2:}\label{Rule 2} If each operation can achieve minimum process energy consumption, the scheduling is more likely to be completed with a relatively minimum energy consumption.

 Based on the above two rules, this paper proposes a heuristic strategy combining total (T) and partial (P) greed. We will take Rule 1 as example to illustrate the combination initialization method. In terms of process time, each operation has its own shortest choice (machine id and speed gear) and the second-shortest choice. By directly combining each operation's shortest choice, we construct a unique particle, denoted as T$_1$. By randomly combining the shortest and second-shortest choice, we generate some particles, denoted as P$_1$. Similarly, in terms of Rule 2, the particles T$_2$ and P$_2$ can be generated. In this paper, Rule 1 generates 40\% of the particles which contain one particle T$_1$ and the rest particles P$_1$. Rule 2 generates 40\% of the particles in the same way as Rule 1. The remaining 20\% of the particles are generated randomly.

\subsection{Update mechanism}
\label{3.5}
\vspace{0.2cm}
This subsection introduces how to organically embed the ideas of DE into PSO's framework and improve its performance. Specifically, DE has three operators: mutation, crossover and selection. The former two can be utilized to improve and maintain population diversity, while the latter one is beneficial to ensure the reasonable convergence. Thus, the mutation and crossover operations are employed in D-DEPSO to generate novel learning exemplar $\widehat{{Pbest}_i}$, which integrate neighbourly information. The selection operator is adopted to guide the population evolutionary direction.
\subsubsection{New learning exemplar}
\vspace{0.2cm}
When using PSO to solve optimization problems, how to select or construct efficient and effective learning exemplars is a crucial work. The traditional multi-objective PSO algorithms randomly select a unique particle in Pareto fronts as the self-guidance $Pbest$. This mechanism is prone to lose high-quantity population diversity, especially in the local search process. Therefore, in this paper, the mutation and crossover ideas of differential evolution algorithm are embedded to enhance information interaction between particles. Note that such information interaction between particles may lead the OS part infeasible. Thus, the following two steps only act on particles' MV part.

\begin{itemize}
\item \textbf{Mutation operation}: As illustrated in the upper right side of Fig.\ref{Fig.flowchart}, for each particle $i~(1\leq i\leq N)$, its own ${Pbest}_i$ and two neighbors’ best experiences ${Pbest}_{i-1}$ and ${Pbest}_{i+1}$ are fused to construct a mutation vector $M_i$:
    \begin{equation}\label{eq.m1}
    M_i = {Pbest}_i\oplus F\cdot({Pbest}_{i-1}\ominus {Pbest}_{i+1}),
    \end{equation}
    where $F$ is scale factor; $i-1$ and $i+1$ represent particle $i$’s two neighbors’ indexes. Specially, particle $1$’s neighbors are particle $N$ and particle $2$, particle $N$’s neighbors are particle $N-1$ and particle $1$. The symbols $\oplus$,$\ominus$ contain two steps. First, the symbol $\ominus$ represents the difference analysis between ${Pbest}_{i-1}$ and ${Pbest}_{i+1}$. If a certain position’s values for ${Pbest}_{i-1}$ and ${Pbest}_{i+1}$ are the same, it means that such value is probably a better choice than the original value in ${Pbest}_{i}$. Therefore, we mark positions with the same values as 0 and positions with different values as 1. Second, the symbol $\oplus$ means to add the ``difference analysis’’ results from the previous step into the initial ${Pbest}_{i}$. That is, in ${Pbest}_{i}$, the positions marked as 1 remain unchanged, while the positions marked as 0 are replaced by ${Pbest}_{i-1}$’s values.

\item \textbf{Crossover operation}: As illustrated in the middle right side of Fig.\ref{Fig.flowchart}, the crossover operation means further information interaction between $M_i$ and the initial $Pbest_i$. In specific, a crossover probability $Cr$ is predefined. For each dimension $d(D\leq d\leq 2\ast D$,$D =\sum_{i=1}^{n}n_i)$ in the MV part, a random number $r$ in the range $[0,1]$ is generated. Then the novel learning exemplar $\widehat{{Pbest}_{i}}$ is calculated as follows:
        \begin{equation}\label{eq.crossover}
    \widehat{{Pbest}_{i,d}} = \left\{\begin{array}{l}
    M_{i,d}, ~if~rand<Cr~or~d~is~equal~to~d', \\
    {Pbest}_{i,d}, ~otherwise.\\
    \end{array}\right.
    \end{equation}
    If $r$ is less than $Cr$, $M_{i,d}$ is retained, otherwise, $M_{i,d}$ is replaced by $Pbest_{i,d}$. In addition, we predefine a fixed dimension $d'$ that directly maintain $M_{i,d'}$ unchanged.

\end{itemize}

Through mutation and crossover operators, the novel $\widehat{{Pbest}_{i}}$ has absorbed more potential and valuable information from other particles, which improves individual quality and enhances population diversity.

\subsubsection{New Learning strategy}
\vspace{0.2cm}
In PSO, particles' transformation from iteration $t$ to $t+1$ is determined by three aspects: inertia, individual cognition and social cognition. The inertia part retains the original information of the previous position of the particle, the individual cognition part is updated based on the historical optimal position of the particle itself, and the social cognition part is updated according to the historical optimal position of the group. In a word, the basic PSO evolution mechanism is based on information fusion of three parts.

\begin{figure}[h]
\begin {center}
\includegraphics[width=2in]{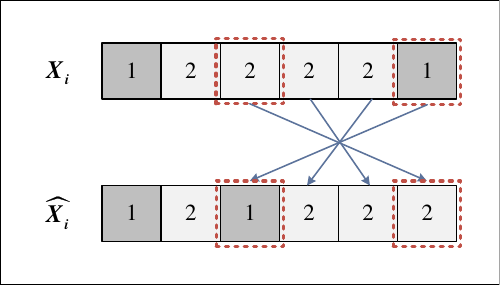}
\caption {The illustration of self-rotation operation.}
 \label{Fig.inertia}
 \end{center}
\end{figure}

On the basis of the above classical framework, and to make PSO more suitable for solving scheduling problems, this paper proposes a new information fusion mechanism: Tripartite Precedence Operation Fusion (TPOF). Specifically, the update formula is given as follows:
\begin{equation}\label{eq.update}
\tilde{X}_i^{t+1} = \omega\cdot\hat{X_i^t} ~\textbf{\circled{\scriptsize{T}}}~ c_1r_1 \cdot\widehat{{Pbest}_i} ~\textbf{\circled{\scriptsize{T}}}~ c_2r_2 \cdot Gbest,
\end{equation}
where $\tilde{X}_i^{t+1}$ is the potential position of iteration $t+1$. $\omega$ represents the weight factor. $c_1$ and $c_2$ are the individual learning factor and the social learning factor respectively. $r_1$ and $r_2$ are the random numbers in [0, 1]. $\hat{X_i^t}$ indicates that a particle's OS part performs self-rotation operation under the action of inertia. As illustrated in Fig.\ref{Fig.inertia}, self-rotation operation refers to the rotation of values between two random points in OS.

In the following, we will provide a detailed introduction of the proposed TPOF's principle and process. For convenience, the formula (\ref{eq.update}) is rewritten as
\begin{equation}\label{eq.TPOF1}
P^* = p_1\cdot P_1~\textbf{\circled{\scriptsize{T}}}~ p_2\cdot P_2~\textbf{\circled{\scriptsize{T}}}~ p_3\cdot P_3.
\end{equation}

Note that the proposed TPOF is based on POX. In POX, two parents participate a crossover operator to generate two children, which can be indicated as $(P_1, P_2) \Rightarrow (C_1, C_2)$. Differently, the proposed TPOF is to generate a new particle $P^*$ obtained by three parents through certain rules, which can be indicated as $(P_1, P_2, P_3)\Rightarrow P^*$. In addition, the canonical POX only considers the OS part, while the proposed TPOF also takes into account the MV part.

\begin{figure}[h]
\begin {center}
\includegraphics[width=4.4in]{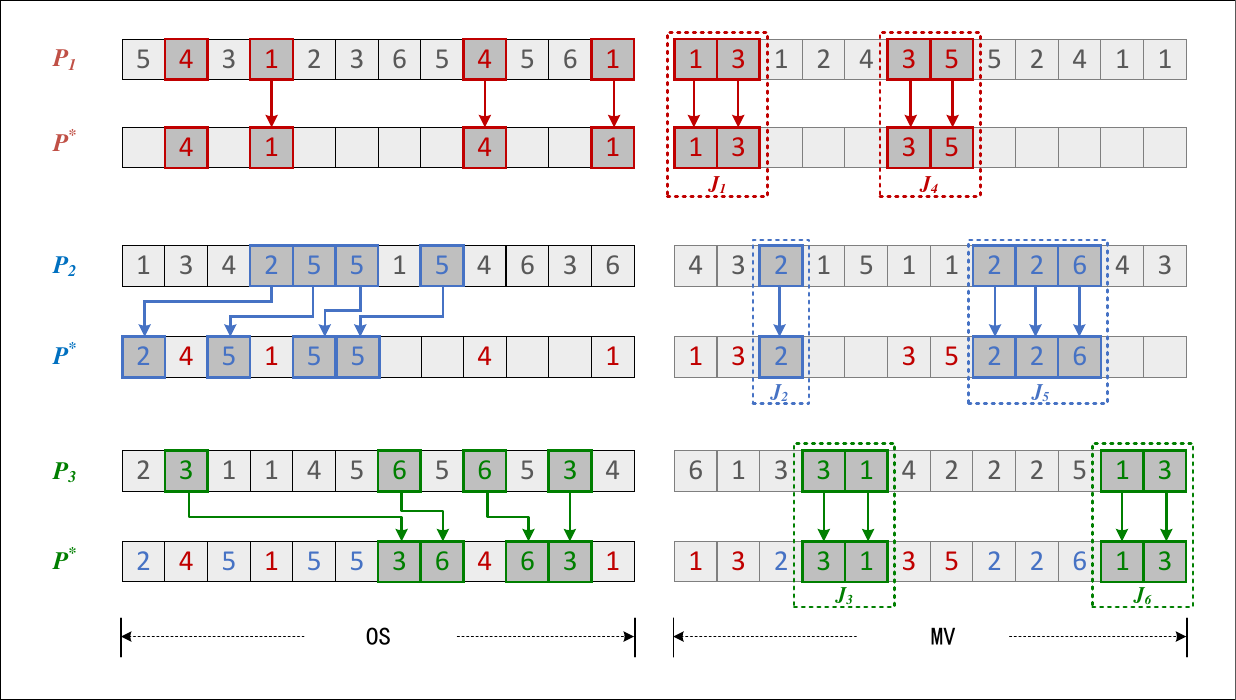}
\caption {The illustration of TPOF.}
 \label{Fig.TPOF}
 \end{center}
\end{figure}

In (38), $p$ is the weight and represents the $P$'s participation degree to the TPOF operation. That is, the higher the weight $p$, the more corresponding $P$'s information we hope the final $P^*$ will contain. To realize such ideation, the subset assignments principle is given as follows:

\begin{equation}\label{eq.TPOF123}
  \left\{\begin{array}{l}
  n_{P_1} = floor({\frac{p_1}{p_1+p_2+p_3}}\cdot{n}), \\[12pt]
  n_{P_2} = floor({\frac{p_1+p_2}{p_1+p_2+p_3}}\cdot{n})-n_{P_1},\\[12pt]
  n_{P_3} = n-n_{P_1}-n_{P_2},\\
  \end{array}\right.
\end{equation}
where $n_{P_1}$ represents the number of jobs in the subset related to $P_1$, and the same applies to $n_{P_2}$ and $n_{P_3}$. Based on the obtained $n_P$, the whole job set is randomly divided into three subsets. To better illustrate the principle of TPOF, we provide a simple example in Fig.\ref{Fig.TPOF}.

As demonstrated in Fig.\ref{Fig.TPOF}, there are 6 jobs in this example. For the sake of simplification, assume that the weights of the three parents are the same. By random allocation, the whole job set is divided into three subsets, $Subset_1$:$~\{J_1,J_4\}$, $Subset_2$:$~\{J_2,J_5\}$, $Subset_3$:$~\{J_3,J_6\}$. First, as for $P_1$'s OS part, all operations of the jobs $J_{1,4}$ are copied to fusion $P^*$. Then, for $P_2$, the operations of $J_{2,5}$ are added into the remaining positions of $P^*$, in order from left to right. Finally, the remaining positions of $P^*$ are filled with the information of $J_{3,6}$ in $P_3$. In terms of MV, vertically merge the information fragments of three subsets into the final $P^*$.

\begin{itemize}
\item \textbf{Selection operation}: Based on (\ref{eq.update}), one can get the potential position of next generation. To ensure that the population evolves towards higher quality, the following selection operation is constructed:
\begin{equation}\label{eq.selection}
X_i^{t+1} = \left\{\begin{array}{l}
\tilde{X}_i^{t+1}, ~~if~\hat{X_i^{t}}~is~dominated~by~\tilde{X}_i^{t+1}, \\[12pt]
\hat{X_i^{t}},~~otherwise.\\
\end{array}\right.
\end{equation}
\end{itemize}

\subsection{Local search}
\label{3.6}
\vspace{0.2cm}
In scheduling problems, the critical path is defined as the longest path from the starting point to the end point, which determines the quality of the final completion time target of the scheduling scheme. Thus, this paper adopts the variable neighborhood search strategy based on the critical path for some well-performed particles. In addition, this work also considers a kind of machine load-based neighborhood structure for energy consumption optimization.
\subsubsection{The critical operation search method considering the setup time} 
\vspace{0.2cm}
The pseudo code for finding critical operations is shown in Algorithm \ref{algorithm2}. The specific steps are as follows:
\begin{enumerate}[label=step\arabic*, ref=step\arabic*]
    \item  In the scheduling table, combine the setup time into the corresponding operations' process time. In this way, the rows' number of the scheduling table is simplified and equals to the total operations' number.
    \item\label{step:2} Find the latest completed operation in the scheduling table, and mark it as a critical operation $CO_\alpha$. Then record its ${Start_{time}}$, ${job}_{id}$, ${operation}_{id}$.
    \item\label{step:3} Search for the previous critical operation of the $CO_\alpha$. The search process is classified into two types:
        \begin{enumerate}
        \item[(i)]Search for the direct precursor in the same job with $CO_\alpha$, denoted as $CO_\beta$. If the search cannot be found, the current operation is the first operation of the corresponding job, then $CO_\beta=0$.
        \item[(ii)]Search for the direct precursor in the same machine with $CO_\alpha$, denoted as $CO_\gamma$. If the search cannot be found, the current operation is the first operation of the corresponding machine, then $CO_\gamma=0$.
        \end{enumerate}
    Compare the completion time of $CO_\beta$ and $CO_\gamma$, in which the one with a later completion time is the previous critical operation of $CO_\alpha$.
    \item If $CO_\beta$ and $CO_\gamma$ are both $0$, then ${Start_{time}}=0$ and the search is over; Otherwise, the operation with a larger value is recorded as the next key operation and the critical path table is updated.
    \item Return to \ref{step:3} and continue until ${Start_{time}}=0$.
\end{enumerate}
\begin{algorithm}[H]
\label{algorithm2}
 \SetAlgoLined
 \KwIn{Schedule}
 \KwOut{Critical path}
 Merges all setup operations in the scheduling into subsequent process operations that are immediately adjacent to it\;
 Find the latest completed operation A\;
 Mark A as a critical operation\;
 Record the completion time of the same job direct precursor J and the same machine direct precursor M for operation A , denoted $CO_\beta$ and $CO_\gamma$\;
 \uIf{$CO_\beta$>$CO_\gamma$}{
  Mark J as a critical operation\;
   }
 \lElse{Mark M as a critical operation}
 \caption{Process of finding critical operations.}
\end{algorithm}

\subsubsection{Variable neighborhood search}
\vspace{0.2cm}
As an improved local search method, variable neighborhood search executes alternately by using the neighborhood structure composed of different actions to achieve better balance between concentration and universality. It is very suitable to be combined with global search algorithms such as PSO to further improve the search efficiency and reconciliation quality. The detailed steps of variable neighborhood search are given as follows:
\begin{enumerate}[label=step\arabic*, ref=step\arabic*]
    \item  Select the top five particles in each generation for local search.
    \item Determine the critical path, and construct multiple neighborhood structures.
        \begin{enumerate}
        \item[(i)]Single point mutation of critical operating machine. An operation is randomly selected in the critical path, whose machine is replaced with the other one among its available machines. The machine speed after replacement is randomly set.
        \item[(ii)]Two points exchange of critical operations. Two operations are randomly selected in the critical path and their positions in the operation sequence are exchanged.
        \item[(iii)]Single point replacement of machines with maximum load. Randomly select an operation in the maximum load machine, and replace its process machine with the other available one. The machine speed after replacement is randomly set.
        \end{enumerate}
    \item Search for new solutions based on the order of step2. If the new solution dominates the current solution, the current solution is updated to the new one. If a better solution cannot be found in the current neighborhood structure, the search continues to the next neighborhood structure until the termination criterion is met. In this paper, the termination criterion is the number of iterations.

\end{enumerate}

\section{Numerical experiments}
\label{Experiment}
\subsection{Construct the EFJSP-M benchmarks}
\label{4.1}
\vspace{0.2cm}
Due to the fact that the EFJSP with machine multi states is newly defined, standard examples are not sufficient for modelling purposes. Therefore, this paper extends the FJSP benchmark MKs\cite{Brandimarte1993RoutingAS} and DPs\cite{brucker1998tabu} to MKA and DPA with reference to the method in \cite{Wei2022HybridES}. The newly generated required data is explained as follows.

Firstly, the single speed is changed to three gears, denoted by $[v_1, v_2, v_3]$, where $v_1$ indicates the minimum speed, $v_3$ indicates the maximum speed as well as the original speed. Correspondingly, the single process time is modified to three notches, i.e., multiplied by $[3, 2, 1]$ respectively on the basis of the original process time. Then, add job setup time as a random integer in $[1, 2]$.

Regarding the different states ($S$) and state switching instructions ($I$), the powers are constructed as follows:
\begin{itemize}
\item[(i)]Power of states. There are mainly four states for an operating machine, namely set up, process, idle and standby. For set up and standby states, the machine power is random number in two intervals $[10, 30]$ and $[3, 5]$, respectively. Note that for the other two states, i.e., process and idle, the machine power will vary with speeds. Thus, by taking $r$ as the basic power, the actual power for the two states can be calculated as the following equation (\ref{eq.E}).
\begin{equation}\label{eq.E}
Power_{process,idle} = r\cdot(2\ast v+1),
\end{equation}
where the range of $r$ are taken as $[30, 50]$ for process state, $[5, 10]$ for idle state respectively.
\item[(ii)]Power of states switching instructions. The energy consumption of the machine to execute the three states switching instructions is also related to the machine speeds. The energy consumption for ‘turn-on’ could be calculated by $(Pprocess$-$Pstandby)$*$Rt$, in which $Rt$ varies uniformly in range $[6, 8]$. By multiplying $(Pprocess$-$Pstandby)$*$Rt$ by $0.2$, one can get the energy consumption for ‘dormancy/dormancy-release’. The energy consumption of ‘turn-off’ is not considered. The energy consumption for ‘acceleration/deceleration’ could be calculated by $Mean(P(v),P(v^{'})) * Rs$, in which $Rs$ varies uniformly in range $[0.2, 0.3]$. By running each basic example 10 times according to the above method, and expanding the original MK01-MK10 to MK01-MK100, the  partial data in the new samples are illustrated in Table \ref{data1} and Table \ref{data2}.
\end{itemize}

\begin{table}[!h]
\centering
\caption{State power consumption of the instances}
\label{data1}
\begin{tabular}{cccccccccc}
\hline
\multirow{3}{*}{Instances} & \multirow{3}{*}{Size} &
\multicolumn{8}{c}{Operating power}\\ \cline{3-10}&
&\multirow{2}{*}{Setup} & \multicolumn{3}{c}{Process}& \multicolumn{3}{c}{Idle} & \multirow{2}{*}{Standby} \\ \cline{4-9}&
& & $v_1$ & $v_2$ &$v_3$ &$v_1$ &$v_2$ &$v_3$ &\\ \hline
MK01&$10\times6$&11.6764&34.5795&69.1591&103.7386&7.5667&15.1334&22.7001&3.3048\\
MK02&$10\times6$&17.9957&35.1974&70.3948&105.5922&7.0003&14.0007&21.0010&3.8628\\
MK16&$10\times6$&24.0220&43.3268&86.6536&129.9803&5.6956&11.3913&17.0869&4.3962\\
MK17&$10\times6$&29.6328&33.1281&66.2562&99.3843&7.2776&14.5552&21.8328&4.2895\\
MK24&$15\times8$&22.2692&41.6450&83.2900&124.9349&5.7037&11.4074&17.1111&4.7399\\
MK25&$15\times8$&18.0916&38.9675&77.9349&116.9024&4.8291&9.6582&14.4872&4.5270\\
MK39&$15\times8$&27.3386&31.7247&63.4494&95.1741&4.8322&9.6644&14.4965&3.7384\\
MK40&$15\times8$&14.7475&40.6174&81.2349&121.8523&3.4575&6.9150&10.3725&3.8106\\
MK45&$15\times8$&25.1865&44.8130&89.6259&134.4389&6.7184&13.4369&20.1553&3.2118\\
MK46&$15\times8$&24.9724&32.4037&64.8075&97.2112&5.6252&11.2505&16.8757&3.6517\\
MK56&$10\times15$&18.6344&49.9512&99.9024&149.8536&7.0580&14.1160&21.1740&3.9713\\
MK57&$10\times15$&26.6630&37.9656&75.9313&113.8969&6.7491&13.4982&20.2473&4.6704\\
MK62&$20\times5$&21.2143&35.3818&70.7637&106.1455&6.7451&13.4902&20.2353&4.0078\\
MK63&$20\times5$&21.5352&30.5171&61.0343&91.5514&5.2327&10.4653&15.6980&4.2926\\
MK74&$20\times10$&21.0546&35.4962&70.9925&106.4887&4.2075&8.4150&12.6225&3.4863\\
MK75&$20\times10$&20.0880&36.9452&73.8905&110.8357&3.4607&6.9215&10.3822&3.2957\\
MK81&$20\times10$&17.1302&31.4649&62.9297&94.3946&5.9550&11.9099&17.8649&4.8204\\
MK82&$20\times10$&29.9096&49.2486&98.4973&147.7459&5.6753&11.3507&17.0260&4.9277\\
MK98&$20\times15$&13.5983&48.5259&97.0518&145.5777&3.3409&6.6818&10.0227&4.1622\\
MK99&$20\times15$&14.1406&45.5006&91.0011&136.5017&7.5709&15.1419&22.7128&4.5651\\ \hline
\end{tabular}
\end{table}

\begin{table}[!h]
\centering
\caption{Instruction switching energy consumption of the instances}
\label{data2}
\vskip 3pt
\begin{tabular}{p{1.5cm}p{1cm}p{1cm}p{1cm}p{1cm}p{1cm}p{1cm}p{1cm}p{1cm}p{1cm}}
\hline
\multirow{3}{*}{Instances} &
\multicolumn{9}{c}{Instruction switching energy consumption}\\ \cline{2-10}
&\multicolumn{3}{c}{Turn on}& \multicolumn{3}{c}{Dormancy/Dormancy-release} &\multicolumn{3}{c}{Acceleration/Deceleration}\\ \cline{2-10}
&$v_1$ & $v_2$ &$v_3$ &$v_1$ &$v_2$ &$v_3$ &$v_1$ &$v_2$ &$v_3$\\ \hline
MK01&30.16&83.71&137.25&6.03&16.74&27.45&3.21&5.35&4.28\\
MK02&19.97&64.51&109.06&3.99&12.90&21.81&3.06&5.09&4.08\\
MK16&8.26&44.46&80.67&1.65&8.89&16.13&2.28&3.80&3.04\\
MK17&19.07&65.51&111.96&3.81&13.10&22.39&2.59&4.32&3.46\\
MK24&6.40&44.25&82.10&1.28&8.85&16.42&1.94&3.23&2.58\\
MK25&2.28&38.71&75.14&0.46&7.74&15.03&1.90&3.17&2.54\\
MK39&7.87&42.64&77.41&1.57&8.53&15.48&1.95&3.24&2.59\\
MK40&2.20&19.32&40.84&0.44&3.86&8.17&1.09&1.82&1.46\\
MK45&24.29&70.82&117.36&4.86&14.16&23.47&2.70 &	4.50 &	3.60\\
MK46&13.42&51.65&89.89&2.68&10.33&17.98 &2.15 &	3.58 &	2.86\\
MK56&19.37&63.66&107.95&3.87&12.73&21.59&3.06 &	5.11 &	4.09\\
MK57&14.77&62.72&110.67&2.95&12.54&22.13&2.35 &	3.92 &	3.13\\
MK62&18.11&62.73&107.35&3.62&12.55&21.47 &2.68 &	4.46 &	3.57\\
MK63&6.34&41.63&76.92&1.27&8.33&15.38&1.98 &	3.30 &	2.64\\
MK74&5.71&39.00&72.29&1.14&7.80&14.46 &1.36 &	2.27 &	1.81\\
MK75&1.21&26.63&52.05&0.24&5.33&10.41 &	1.14 &	1.90 &	1.52\\
MK81&7.79&48.67&89.55&1.56&9.73&17.91 &	1.96 &	3.27 &	2.61\\
MK82&4.56&39.20&73.83&0.91&7.84&14.77 &	1.80 &	3.00 &	2.40\\
MK92&6.00&18.40&42.80&1.20&3.68&8.56 &	1.32 &	2.20 &	1.76\\
MK99&18.95&66.67&114.40&3.79&13.33&22.88& 	2.61& 	4.34 &	3.48\\ \hline

\end{tabular}
\end{table}

\subsection{Performance Evaluation Criteria}
\label{4.2}
\vspace{0.2cm}
The following three performance indicators commonly used for multi-objective algorithms are adopted to evaluate the proposed D-DEPSO:
\begin{enumerate}
\item[(1)]Inverse generational distance (IGD)

IGD mainly evaluates the convergence and distribution of the algorithm, i.e., it represents the mean of the shortest distance between the reference set $P^*$ and the Pareto fronts of solution set $A$. The reference set $P^*$ is the Pareto fronts of whole solutions abstained by all algorithms. IGD is calculated as follows:
\begin{equation}\label{eq.evaluation2}
IGD(A,P^{*}) = \frac{\sum_{x\in P^{*}}d(x,PF)}{|P^{*}|},PF\subseteq A,
\end{equation}
where $PF\subseteq{A}$ represents the Pareto fronts of set $A$. $|P^{*} |$ represents the number of solutions in $P^{*}$. The smaller the IGD, the better the solution in $A$.
\item[(2)]Hypervolume (HV)

In multi-objective optimization problems, hypervolume refers to a hypercube surrounded by $A$'s Pareto fronts and a reference point $r$ (usually an ideal point or a bad point). HV reflects the distribution and convergence of the non-dominated solution set in the target space. HV is calculated as follows:
\begin{equation}\label{eq.evaluation3}
HV(A,r) = \bigcup_{a\in PF} v(a,r),
\end{equation}
where $v(a,r)$ represents the super-volume of $A$'s Pareto fronts and the reference point $r$. The larger the HV, the better the solution in $A$.
\item[(3)]Coverage metric (C-metric)

C-metric describes the dominant relationship between two solution sets $A$ and $B$. In specific, the symbol $C(A,B)$ represent the percentage of B's solutions which are dominated by at least one solution in A. It is calculated as follows:
\begin{equation}\label{eq.evaluation3}
C(A,B) = \frac{b\in B|\exists a\in A:a\prec b}{|B|},
\end{equation}
where $C(A,B)=1$ means the solutions in $B$ are all dominated by the solutions in $A$. $C(A,B)=0$ means that none of the solutions in B are dominated by the solutions in $A$. The larger the C-metric, the better the solution in $A$.
\end{enumerate}

\subsection{Parameter settings}
\label{4.3}
\vspace{0.2cm}
For the D-DEPSO, six parameters determine the algorithm performance, namely the population size ($N$), inertia weight factor ($\omega$), individual cognitive factor ($r_1$), social cognitive factor ($r_2$), scaling factor ($F$) and crossover probability ($Cr$). The existing research suggests that dynamic inertia weights and cognitive factors in PSO can better achieve the balance between exploration and exploitation. In this work, $\omega$, $r_1$, $r_2$ are determined by the following formulas:
\begin{equation}\label{eq.setting1}
\omega = 2-\frac{iter\cdot(2-0.4)}{max\_iter},
\end{equation}
\begin{equation}\label{eq.setting2}
c_1 = 2-\frac{iter\cdot(2-1.5)}{max\_iter\cdot rand()},
\end{equation}
\begin{equation}\label{eq.setting3}
c_2 = 1.5+\frac{iter\cdot(2-1.5)}{max\_iter\cdot rand()}.
\end{equation}

\begin{figure}[h]
\begin {center}
\includegraphics[height=1.8in,width=5in]{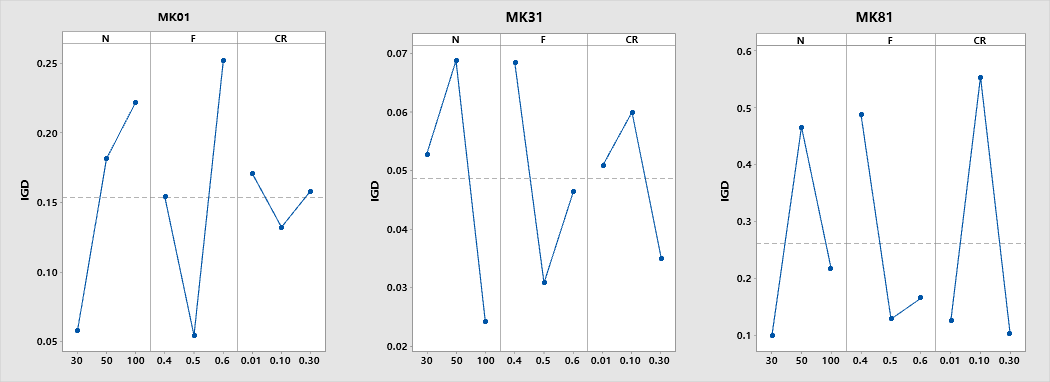}
\caption {Main effects plot for means of IGD.}
 \label{Fig.factor}
 \end{center}
\end{figure}

The Taguchi method is used to obtain the optimal combination of the remaining parameters. Main parameters and factor settings are reported in Table \ref{yinziShuipingbiao}. In addition, Table \ref{yinziShuipingbiao} also shows the IGD values of D-DEPSO for some examples under each parameter combination. Based on the above data, the factor level trend chart of key parameters is drawn in Fig.\ref{Fig.factor}. According to the experimental results, one can get the optimal parameter combination setting: $N = 30$, $F = 0.5$, $Cr = 0.3$.
\begin{table}[!h]
\centering
\caption{Factor level table and IGD of MK01,31,81.}
\label{yinziShuipingbiao}
\vskip 3pt
\begin{tabular}{ccccccc}
\hline
  $Combinations$ & $N$ & $F$ & $Cr$ & $MK01$& $MK31$& $MK81$\\
\hline
 $1$ & 30 & 0.4 & 0.01 & 0.0250 & 0.0798 & 0.1076\\
 $2$ & 30 & 0.5 & 0.10 & 0.0771 & 0.0477 & 0.1356\\
 $3$ & 30 & 0.6 & 0.30 & 0.0727 & 0.0724 & 0.0550\\
 $4$ & 50 & 0.4 & 0.10 & 0.0727 & 0.0554 & 1.1963\\
 $5$ & 50 & 0.5 & 0.30 & 0.0361 & 0.1026 & 0.0928\\
 $6$ & 50 & 0.6 & 0.01 & 0.4360 & 0.1614 & 0.1121\\
 $7$ & 100 & 0.4 & 0.30 & 0.3667 & 0.0168 & 0.1612\\
 $8$ & 100 & 0.5 & 0.01 & 0.0514 & 0.0041 & 0.1609\\
 $9$ & 100 & 0.6 & 0.10 & 0.2478 & 0.0399 & 0.3325\\
\hline
\end{tabular}
\end{table}

\subsection{Verify the EFJSP-M model}
\label{4.4}
\vspace{0.2cm}
In order to verify the validity of the proposed EFJSP-M, we employed data set DPA, in which five samples will be solved by the IBM ILOG CPLEX 12.10.0.0 optimization solver within the maximum running time 1800 seconds. In addition, the corresponding energy consumption will be also calculated. As comparison, the proposed D-DEPSO algorithm is adopted to solve the same examples with the maximum iterations number 300. The comparison results between CPLEX and D-DEPSO are listed in Table \ref{moxingjyanzheng}, in which the timeout results are marked by $'-'$. The summary of Table\ref{moxingjyanzheng} is given as follows:

        \begin{enumerate}
        \item[(i)] For CPLEX, it can obtain the optimal $C_{max}$ in the small-scale cases DPA01-03 within two seconds, but can not reach the feasible solution in the large-scale samples DPA04-05 within the set time. In addition, the iterations number and running time of CPLEX increase exponentially with the increase of the problem scale. This indicates that the proposed EFJSP-M model is theoretically reasonable, but there are certain difficulties in the technical implementation of finding the optimal solution.
        \item[(ii)] For D-DEPSO, it can find the corresponding $C_{max}$ on all five cases DPA01-05. Among them, it obtains the same optimal solution as CPLEX in case DPA01. As for DPA02-03, the solutions obtained by D-DEPSO are close to CPLEX's optimal solutions, and with lower energy consumptions. In addition, the iterations number and running time of D-DEPSO vary linearly with the increase of problem scale. These suggest that D-DEPSO demonstrates superior calculation ability and robustness for solving the proposed EFJSP-M model.
\end{enumerate}

\begin{table}[!h]
\centering
\caption{Calculation results of CPLEX and D-DEPSO.}
\label{moxingjyanzheng}
\vskip 3pt
\begin{tabular}{cccccc|cccc}
\hline
\multirow{2}{*}{$Instances$} & $Size$ &
\multicolumn{4}{c}{$CPLEX$}&\multicolumn{4}{c}{$D-DEPSO$}\\ \cline{3-10}
 & $n\times n_i\times m$ & $C_{max}$ & $TEC$ & $Iterations$ & $CPU(s)$ & $C_{max}$ & $TEC$ & $Iterations$ & $CPU(s)$ \\ \hline
 DPA01 & $2\times 3\times 3$ &  \textbf{158} & \textbf{36643} & \textbf{43} & \textbf{0.387} & \textbf{158} & \textbf{36643} & 300 & 8.42 \\
 DPA02 & $3\times 5\times 4$ & \textbf{288} & 90112 & \textbf{280} & \textbf{0.754} & 296 & \textbf{86564} & 300 & 12.9 \\
 DPA03 & $4\times 7\times 5$ & \textbf{450} & 17208 & 5730 & \textbf{1.497} & 476 & \textbf{168413} & \textbf{300} & 27.08 \\
 DPA04 & $8\times 10\times 4$ & $-$ & $-$ & $-$ & 1800 & \textbf{1116} & \textbf{427453} & \textbf{300} & \textbf{112.78} \\
 DPA05 & $10\times 15\times 5$ & $-$ & $-$ & $-$ & 1800 & \textbf{1745} & \textbf{880655} & \textbf{300} & \textbf{217.99} \\ \hline
\end{tabular}
\end{table}

\subsection{Compare with D-DEPSO variants}
\label{4.5}
\vspace{0.2cm}
The proposed D-DEPSO has several components, such as initialization mechanism, DE operators and critial path local search method. In order to verify the effectiveness of these components, a comparative experiment was designed in this section. The experiment includes the proposed algorithm D-DEPSO, non-hybrid initialization mechanism (NHI), non-DE method (NDE) and non-critical path local search method (NCP). The comparison experiment is designed based on three considerations. Firstly, particle swarm optimization relies on initial population, and hybrid initialization method can reduce the blindness of particle search. Secondly, DE operator can improve the quality of learning samples, and then affect the efficiency of algorithm search. Finally, the local search method based on the critical path is helpful to further improve the accuracy of the algorithm.

\begin{figure}[h!t]
\begin {center}
\includegraphics[height=3.4in,width=6in]{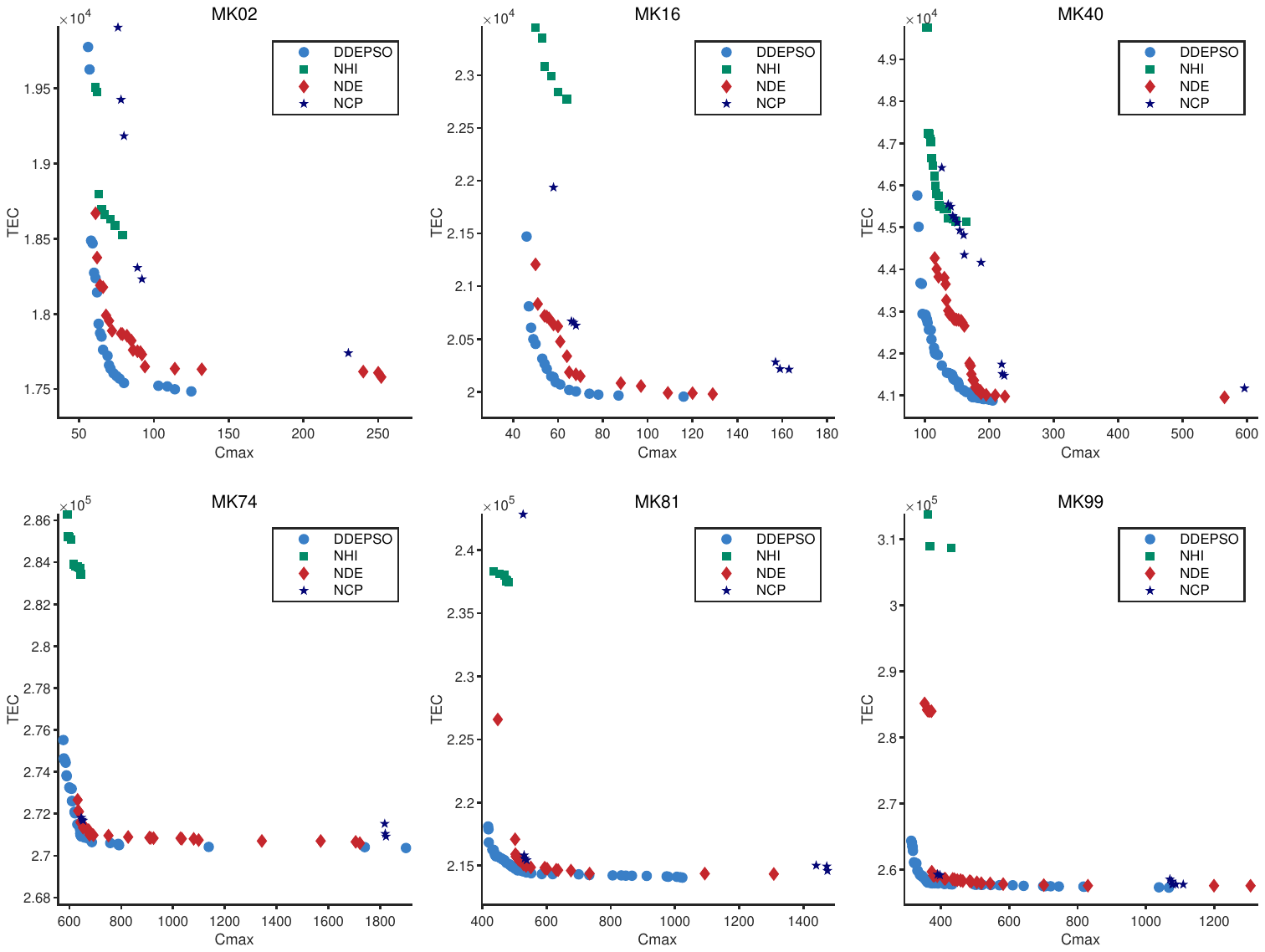}
\caption {The Pareto fronts scatter plot obtained by running D-DEPSO and its variants.}
 \label{Fig.PF1}
 \end{center}
\end{figure}

In Fig.\ref{Fig.PF1}, the Pareto fronts scatter plot obtained by D-DEPSO and its reduced variants on samples MK02, MK16, MK40, MK74, MK81 and MK99 are presented. As can be seen from Fig.\ref{Fig.PF1}, the performance of D-DEPSO is significantly better than that of NHI, which is due to the reason that in large scale of flexible job-shop scheduling problems, the heuristic strategy can reduce the blindness of the algorithm search and help the algorithm quickly converge to the optimal solution. As for the NDE, D-DEPSO is obviously superior on small-scale samples MK02,16,40. For large-scale samples MK74,81,99, the Pareto fronts of D-DEPSO have better universality than NDE. Therefore, DE can indeed improve the quality of learning samples and enhance the local search ability of the algorithm. For NCP, its solutions have poor performance in various cases than those of D-DEPSO, and the number of Pareto solutions is small and scattered. This suggests that the variable neighborhood search method based on critical path has strong local search ability to help PSO jumping out of the local optimal. To sum up, each component in D-DEPSO improves the performance of the algorithm to a certain extent.

\begin{figure}[h!t]
\begin {center}
\includegraphics[height=3.5in,width=6.5in]{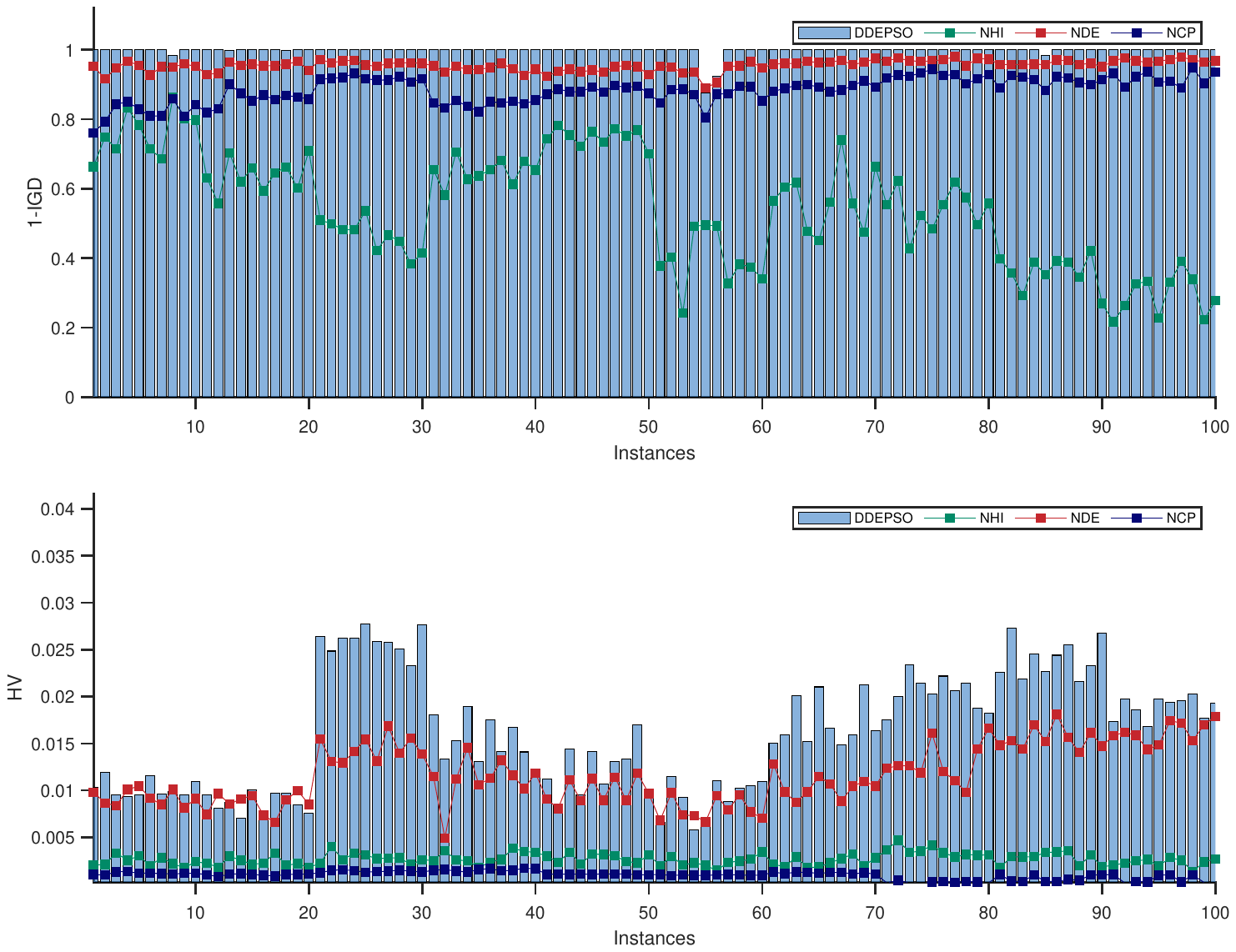}
\caption {IGD and HV comparison of D-DEPSO and its variants on MK01-MK100.}
 \label{Fig.IGD_HV1}
 \end{center}
\end{figure}

\begin{figure}[h!t]
\begin {center}
\includegraphics[height=3.2in,width=6in]{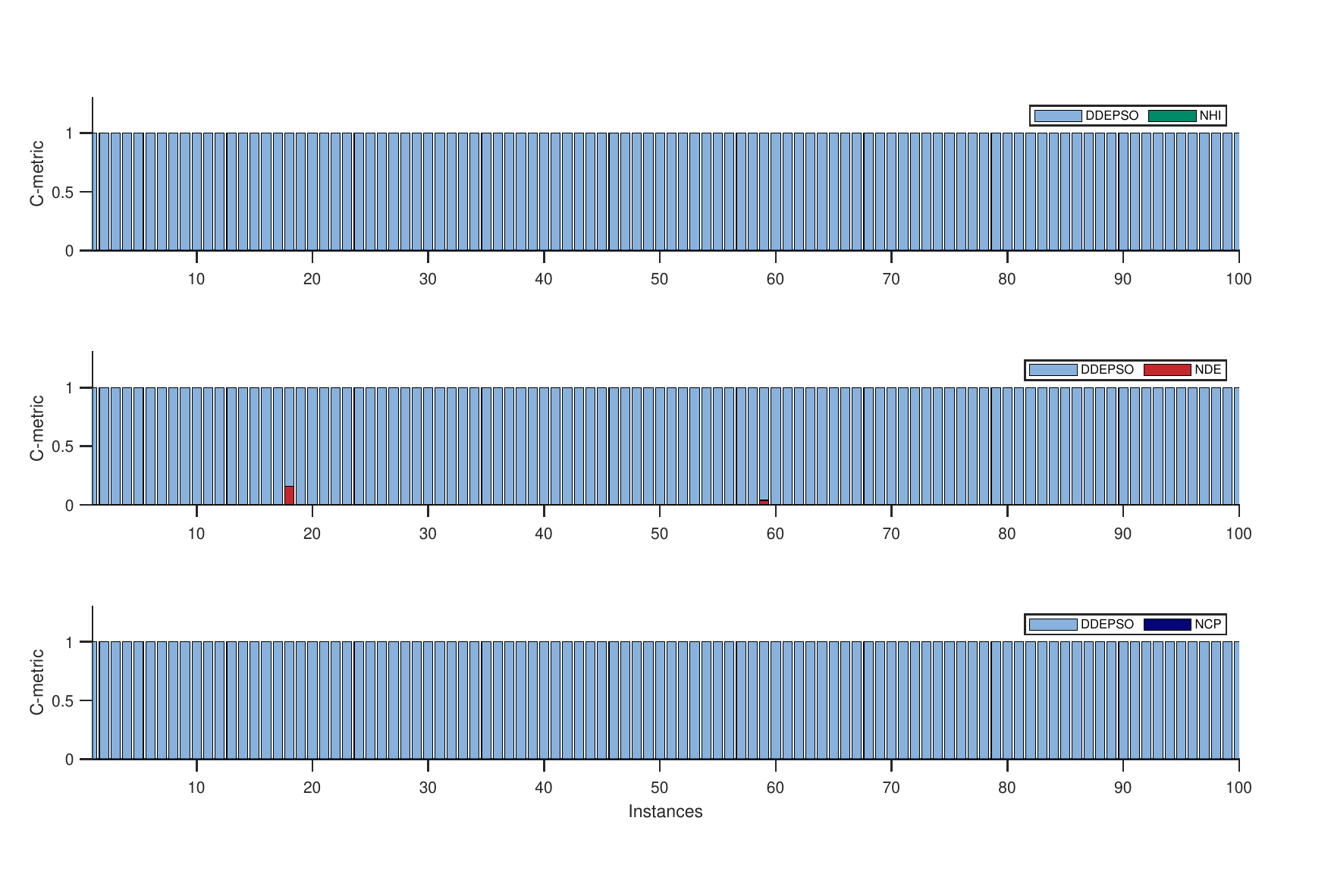}
\caption {C-metric comparison of D-DEPSO and its variants on MK01-MK100.}
 \label{Fig.C1}
 \end{center}
\end{figure}

The experimental results based on the IGD and HV are shown in Fig.\ref{Fig.IGD_HV1}. To make the observation clearer, we take 1-IGD instead of the original IGD value. The larger the value, the closer the Pareto solution set obtained by the algorithm is to the optimal solution. The results of C-metric are shown in Fig.\ref{Fig.C1}. The length of the bar represents the size of the C-metric, and a larger value indicates that the current algorithm has a higher solution set coverage than its comparison algorithm.

According to Fig.\ref{Fig.PF1}, Fig.\ref{Fig.IGD_HV1} and Fig.\ref{Fig.C1}, D-DEPSO is significantly higher than other variants except MK55 in IGD index. The HV index of D-DEPSO is partially lower than NDE in some small-scale samples, but is significantly better than NDE on the Pareto fronts. This suggests that D-DEPSO can achieve better non-dominated solutions, which have slightly concentrated distribution. As for large-scale samples, the HV index of D-DEPSO is significantly better. The Pareto solutions obtained by other variants are always dominated by one of the Pareto solutions in D-DEPSO.

\subsection{Compare with other algorithms}
\label{4.6}
\vspace{0.2cm}
To comprehensively verify the effectiveness of the proposed D-DEPSO, five other popular algorithms are employed in the following comparison experiments. Among five algorithms, two currently proposed particle swarm optimization algorithms are adopted, that is, hybrid quantum particle swarm optimization (QPSO) \cite{Xu2024HybridQP} and two-stage multi-objective particle swarm optimization algorithm (MOPSO) \cite{Shi2023ANB}. The other three excellent algorithms are hybrid shuffled frog-leaping algorithm (HSFLA) \cite{Meng2023MILPMA}, hybrid genetic algorithm (HGA) \cite{Sun2023HybridGA} and NSGA-II \cite{Wei2022HybridES}. The specific parameters of the corresponding algorithms are set by referring to the original text, and the particles number and iteration times are the same as those of D-DEPSO. All algorithms are independently run 10 times on all samples.

\begin{figure}[h!t]
\begin {center}
\includegraphics[height=3.4in,width=6in]{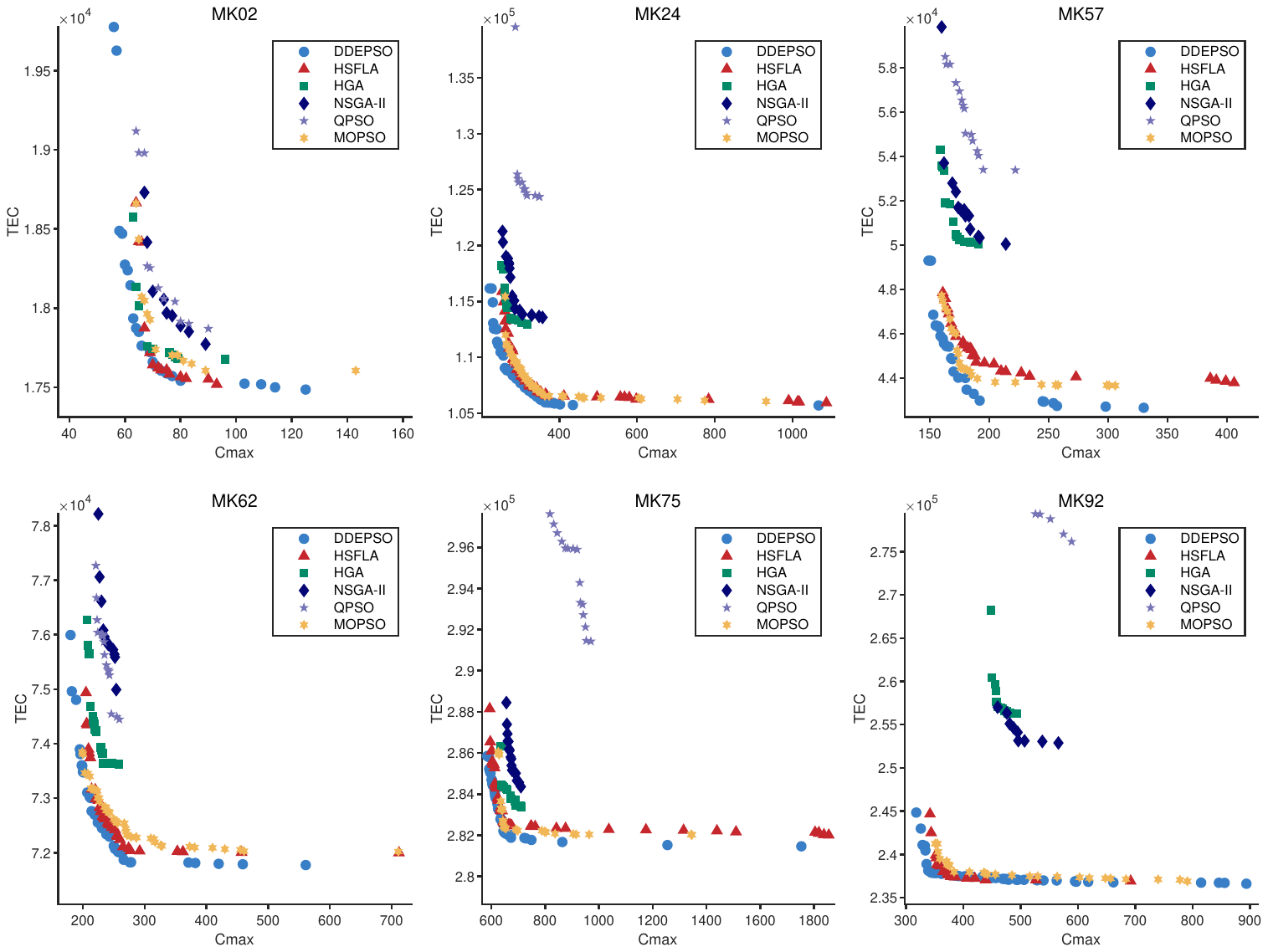}
\caption {The Pareto fronts scatter plot obtained by running D-DEPSO and five other algorithms.}
 \label{Fig.PF2}
 \end{center}
\end{figure}

\begin{figure}[h!t]
\begin {center}
\includegraphics[height=3.5in,width=6.5in]{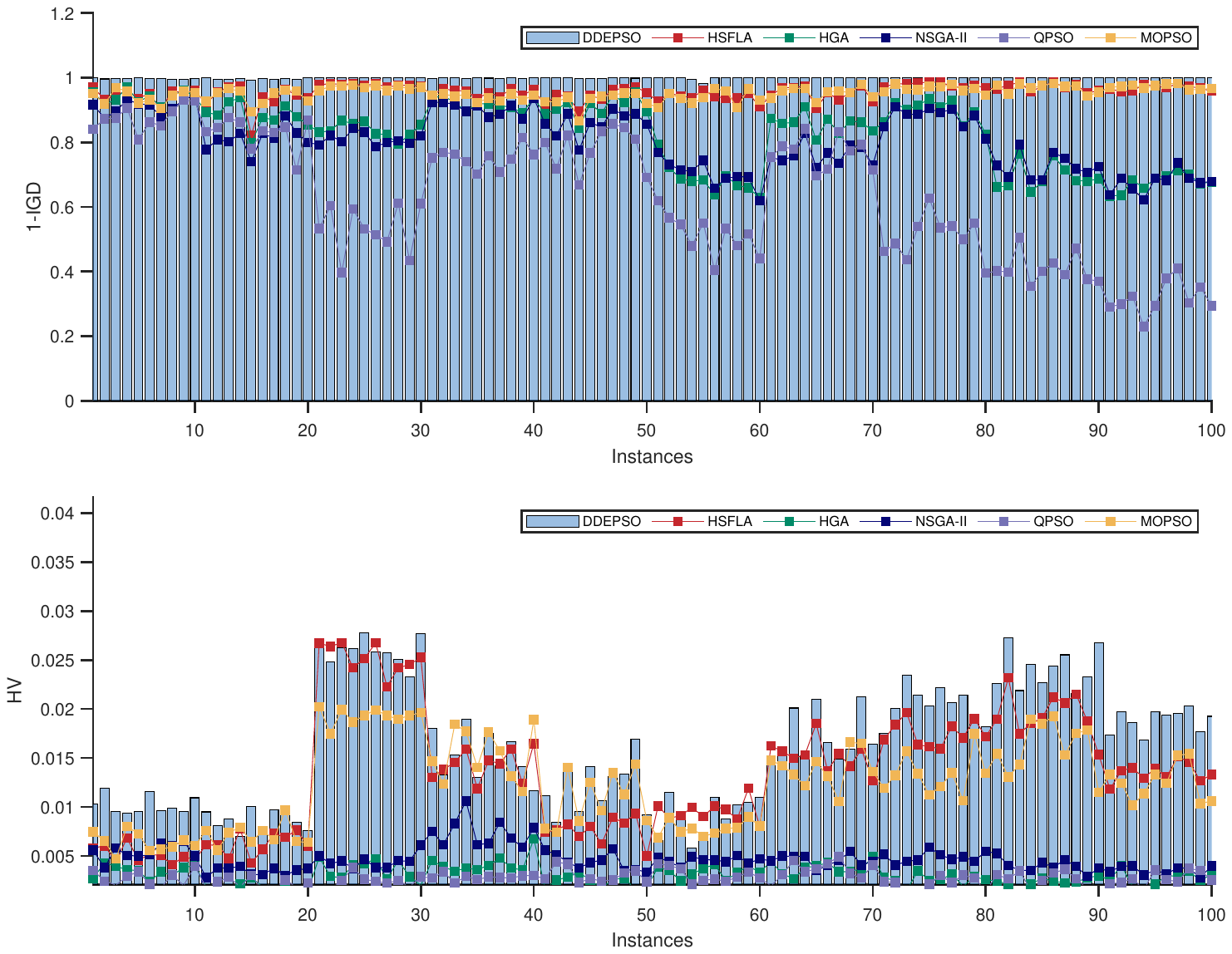}
\caption {IGD and HV comparison of D-DEPSO and five other algorithms on MK01-MK100.}
 \label{Fig.IGD_HV2} 
 \end{center}
\end{figure}

\begin{figure}[h!t]
\begin {center}
\includegraphics[height=3.2in,width=6in]{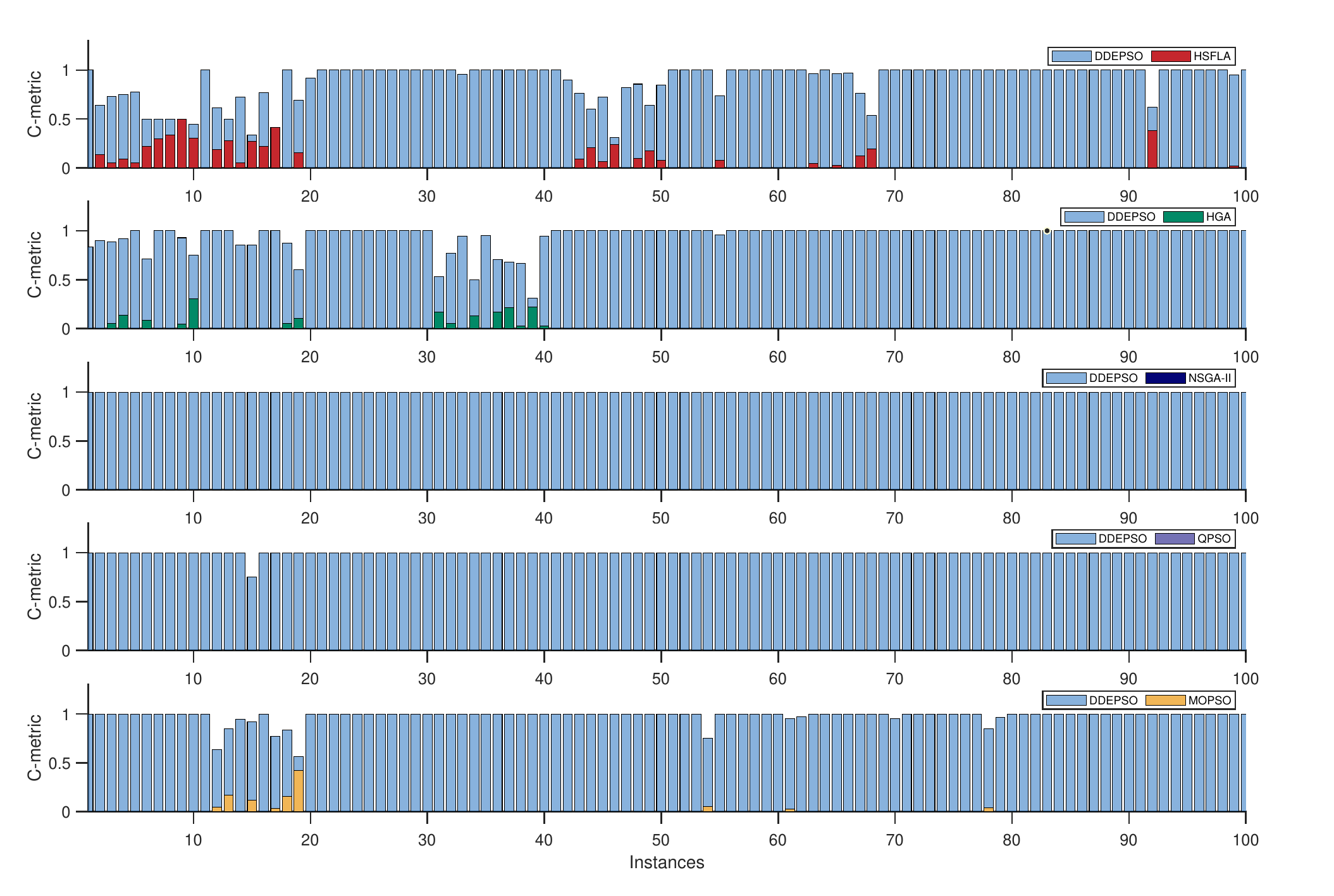}
\caption {C-metric comparison of D-DEPSO and five other algorithms on MK01-MK100.}
 \label{Fig.C2}
 \end{center}
\end{figure}

To intuitively display the optimization performances of algorithms, several samples with different scales are chosen, including MK02, MK24, MK57, MK62, MK75, MK92. The experiments results conclude the Pareto fronts in Fig.\ref{Fig.PF2}, IGD and HV results in Fig.\ref{Fig.IGD_HV2}, and C-metric results in Fig.\ref{Fig.C2}.

As can be seen from Fig.\ref{Fig.PF2}, in the small-scale example MK02, the Pareto fronts of D-DEPSO and HSFLA are closer to the origin, that is, both show good convergence. In the other examples, D-DEPSO demonstrates the best convergence results. In addition, the distribution of the Pareto fronts of D-DEPSO in all examples is more uniform, which suggests that it could search effectively in a broader scope.

From Fig.\ref{Fig.IGD_HV2}, D-DEPSO obtains the optimal values in all samples on IGD. The MOPSO and HSFLA algorithms also have good performance, ranking second and third places. As for HV, the larger HV value represents the better diversity of the solution set of the algorithm. D-DEPSO has a few samples slightly worse than HSFLA on MK21-MK30 and MK51-MK60, and a few samples slightly worse than MOPSO on MK31-MK40. For over 75\% of the examples, the proposed D-DEPSO has the optimal HV value.

From Fig.\ref{Fig.C2}, compared with HSFLA, the proposed D-DEPSO performs slightly worse on MK09, but better on the rest small-scale examples. On medium and large-scale examples, C values are close to 1 on more than 90\% samples. Compare with HGA and MOPSO, DDEPSO performs better on all examples, and almost all C values are close to 1 on medium and large scale examples. Compared with NSGA-II and QPSO, D-DEPSO has C=1 on all examples, which means complete coverage.

\begin{figure}[h!t]
\begin {center}
\includegraphics[width=6in]{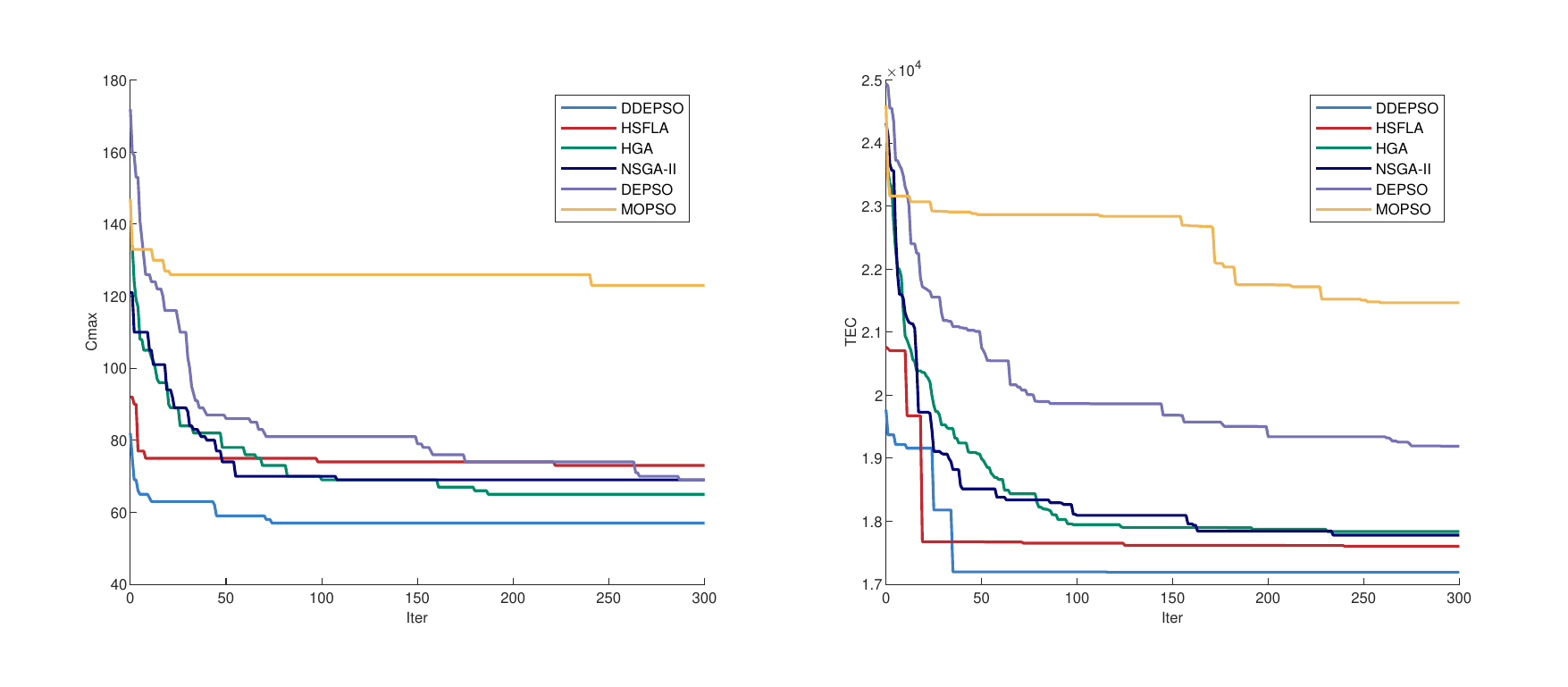}
\caption {The convergence diagram of two targets for each algorithm on sample MK02.}
 \label{Fig.fitness}
 \end{center}
\end{figure}

Fig.\ref{Fig.fitness} shows the convergence curves of example MK02. From Fig.\ref{Fig.fitness}, it can be observed that D-DEPSO and HSFLA can decrease rapidly in the early iterations. Moreover, D-DEPSO can find the optimal solution for both goals. It is worth noting that the two objectives of the above two algorithms are significantly better than other algorithms in the initial iteration. This is because that they both employ the heuristic initialization strategy, while other algorithms use a random initialization strategy or a chaotic initialization strategy. This indicates that the heuristic initialization strategy is indeed beneficial to improve algorithm's performance in FJSP.

\begin{figure}[h!t]
\begin {center}
\includegraphics[width=6.5in]{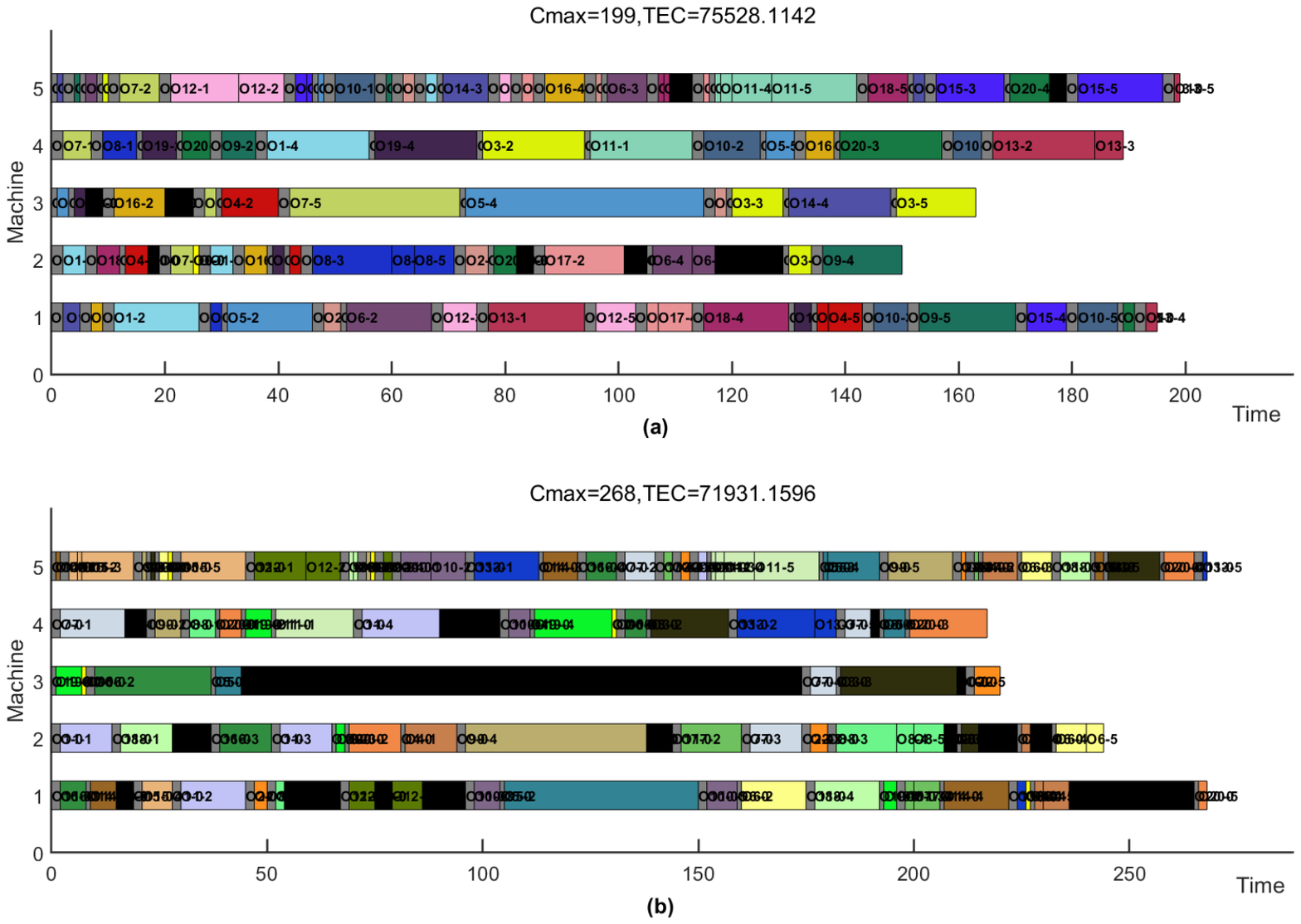}
\caption {The Gantt Diagram of D-DEPSO's solution on sample MK02:(a)min Cmax,(b)min TEC.}
 \label{Fig.gant}
 \end{center}
\end{figure}

The Gantt Diagram of D-DEPSO's solutions on sample MK02 is shown in Fig.\ref{Fig.gant}. In specific, Fig.\ref{Fig.gant}(a) is the Gantt Diagram with shortest completion time among all the Pareto solutions, and Fig.\ref{Fig.gant}(b) is the one with the lowest energy consumption among all the Pareto solutions. Each rectangular block in the figure is an operation, and rectangles of the same color belong to a same job. The gray rectangular block indicates the ready state and the black rectangular block indicates the idle/standby states.

\section{Conclusion and disscussion}
\label{Conclusion}
In this paper, we proposed a novel EFJSP-M considering machine multi states, which includes multi speeds and setup time. To address the proposed problem, a kind of D-DEPSO was designed. The proposed D-DEPSO contains the following three aspects to improve optimization performance: a hybrid initialization strategy, an updating mechanism embedded with differential evolution operators and a critical path variable neighborhood search strategy. Finally, based on the DPs dataset, the feasibility of the model and the effectiveness of D-DEPSO were verified using CPLEX. Based on the MKs dataset, a comparison was made with five advanced algorithms to further validate the efficiency of D-DEPSO from the perspectives of Pareto fronts, inverse generation distance, hypervolume, and solution coverage.

It is worth noting that the hybrid initialization method designed in this paper demonstrates superior performance in the EFJSP-M. Naturally, in the future research, it is of interest to consider whether it also has excellent guidance effectiveness in other schedule problems and other algorithms. In addition, the machine multi states considered in this paper conforms to the real production process and has certain guiding significance for practical production. Considering dynamic scenarios such as order arrival, machine failure in real situations, we will further study the complex machine states in dynamic environments in the future to better adapt to real-life production activities.

\section*{Acknowledgements}
The research is supported by National Natural Science Foundation of China (No.62172262), the Program of Shandong Provincial Natural Science Foundation (No.ZR2022MF318, No.ZR2020MA055, and ZR2020MG003).

\printcredits
\bibliographystyle{elsarticle-num}

\bibliography{cas-refs}
%
%

\end{document}